\documentclass[a4paper,fleqn]{cas-dc}
\usepackage[table]{xcolor}
\usepackage{tabularx}
\definecolor{headerblue}{RGB}{225,238,250}
\definecolor{lastgray}{RGB}{242,242,242}
\usepackage[authoryear,round]{natbib}
\usepackage{xcolor}
\usepackage{hyperref}

\hypersetup{
    colorlinks=true,
    citecolor=blue
}

\usepackage{amsmath}
\usepackage{amssymb}

\usepackage{graphicx}
\usepackage{booktabs}
\usepackage{array}

\usepackage{ragged2e}
\usepackage{microtype}
\ExplSyntaxOn

\cs_set:Npn \__make_fig_caption:nn #1#2
{
    \skip_vertical:N \l_fig_abovecap_skip

    \parbox{\l_fig_width_dim}
    {
        \normalfont\normalsize
        \justifying
        \setlength{\parindent}{0pt}
        \noindent
        \textbf{\color{scolor}#1:}~#2\par
    }

    \skip_vertical:N \l_fig_belowcap_skip
}

\cs_set:Npn \__make_tbl_caption:nn #1#2
{
    \skip_vertical:N \l_tbl_abovecap_skip

    \parbox{\l_tbl_width_dim}
    {
        \normalfont\normalsize
        \justifying
        \setlength{\parindent}{0pt}
        \noindent
        \textbf{\color{scolor}#1:}~#2\par
    }

    \skip_vertical:N \l_tbl_belowcap_skip
}

\ExplSyntaxOff

\begin{document}

\let\WriteBookmarks\relax
\def\floatpagepagefraction{1}
\def\textpagefraction{.001}

\shorttitle{Reliability-Aware Multi-Strategy Guidance Network}
\shortauthors{Zhao et al.}


\title[mode=title]{
Thermal-Guided Low-Light UAV Image Enhancement: A Reliability-Aware Guidance Network and a Unified Benchmark
}


\author[1,2]{Zhicheng Zhao}

\author[1,3]{Yu Wang}

\author[3,4]{Jiangong Xu}
\cormark[1]
\ead{dd_xjg@ahu.edu.cn}

\author[5]{Lixiang Xu}

\author[1,6]{Chenglong Li}

\author[3,4]{Jin Tang}


\affiliation[1]{
    organization={School of Artificial Intelligence, Anhui University},
    city={Hefei},
    postcode={230601},
    country={China}
}

\affiliation[2]{
    organization={Anhui Provincial Key Laboratory of Security Artificial Intelligence, Anhui University},
    city={Hefei},
    postcode={230601},
    country={China}
}

\affiliation[3]{
    organization={Anhui Provincial Key Laboratory of Multimodal Cognitive Computation, Anhui University},
    city={Hefei},
    postcode={230601},
    country={China}
}

\affiliation[4]{
    organization={School of Computer Science and Technology, Anhui University},
    city={Hefei},
    postcode={230601},
    country={China}
}

\affiliation[5]{
    organization={School of Artificial Intelligence Engineering, Hefei Institute of Technology},
    city={Hefei},
    postcode={230601},
    country={China}
}

\affiliation[6]{
    organization={State Key Laboratory of Opto-Electronic Information Acquisition and Protection Technology, Anhui University},
    city={Hefei},
    postcode={230601},
    country={China}
}

\cortext[cor1]{Corresponding author.}

\begin{abstract}
Thermal-guided low-light image enhancement (LLIE) is promising for nighttime unmanned aerial vehicle (UAV) perception because thermal imagery remains informative under severe visible degradation. However, the degraded visible image serves as both the alignment reference and the enhancement target, making thermal guidance uncertain in its spatial correspondence and local utility under low illumination, sensor differences, and UAV motion. Misaligned or unsuitable thermal cues may introduce artifacts, while spatially varying degradation requires region-specific restoration rather than uniform enhancement. To address these challenges, we propose a Reliability-Aware Multi-Strategy Guidance Network (RMGNet) for thermal-guided UAV LLIE under weak optical-thermal alignment. Its Reliability-Aware Guidance Calibration Module (RAGCM) performs coarse-to-fine displacement estimation and jointly evaluates the reliability of aligned thermal guidance and the usability of degraded visible evidence. Based on calibrated guidance, degradation, and reliability cues, the Multi-Strategy Enhancement Routing Module (MSERM) generates candidates for detail preservation, highlight-protected relighting, and thermal compensation, then spatially routes their contributions. We further construct the UAV weakly aligned Optical-Thermal dataset (UWOT), the largest UAV-oriented benchmark to date for weakly aligned optical-thermal low-light enhancement, containing 5,000 triplets. Extensive experiments show that RMGNet outperforms representative single-modal and multimodal state-of-the-art methods in enhancement quality while retaining competitive computational efficiency and greater robustness to practical cross-modal misalignment. The dataset and code will be publicly available at \url{https://github.com/mmic-lcl/Datasets-and-benchmark-code}.
\end{abstract}




\maketitle

\section{Introduction}

Unmanned aerial vehicles (UAVs) support nighttime surveillance, infrastructure inspection, search and rescue, and navigation. However, limited illumination, sensor noise, and platform motion degrade visible images through low contrast, color distortion, blur, and detail loss \citep{xu2022snr,darkir}. These degradations impair visual quality and obscure the structural and semantic cues required by downstream perception, particularly for small objects and narrow boundaries. Reliable low-light image enhancement (LLIE) is therefore important for nighttime UAV vision \citep{Retinexformer}.

Visible-only LLIE methods have made substantial progress in illumination adjustment, noise suppression, color recovery, and joint degradation modeling \citep{Chen2018Retinex,xu2022snr,Retinexformer,hvi,darkir}. Nevertheless, their effectiveness remains constrained by the information retained in the visible observation. In severely underexposed UAV imagery, object boundaries and small structures may be weakly represented or obscured by noise. Their restoration then relies heavily on learned image priors, and multiple plausible reconstructions may be compatible with the same degraded input \citep{wang2022llflow,VIIS}. Complementary observations can provide additional evidence for constraining these ambiguous structures.

Thermal imagery is less dependent on visible illumination and can provide complementary contours when visible observations are severely degraded \citep{cao2023deep,wang2025thermal}. However, differences in sensor geometry, spatial resolution, and field of view, together with parallax and platform motion, can introduce residual spatial discrepancies between UAV visible and thermal images \citep{chen2024oafa}. Here, weak alignment refers to scene-level correspondence accompanied by local spatial discrepancies. Transferring displaced thermal structures into the visible image may produce shifted boundaries, duplicated contours, or blurred details. Low-light degradation further weakens the visible structures used to estimate correspondence, particularly in regions where auxiliary information is most needed. The first challenge is therefore to establish locally reliable optical–thermal correspondence and assess the reliability of thermal guidance under degraded visible observations.

\begin{figure}[pos=t]
    \centering
    \includegraphics[width=\columnwidth]{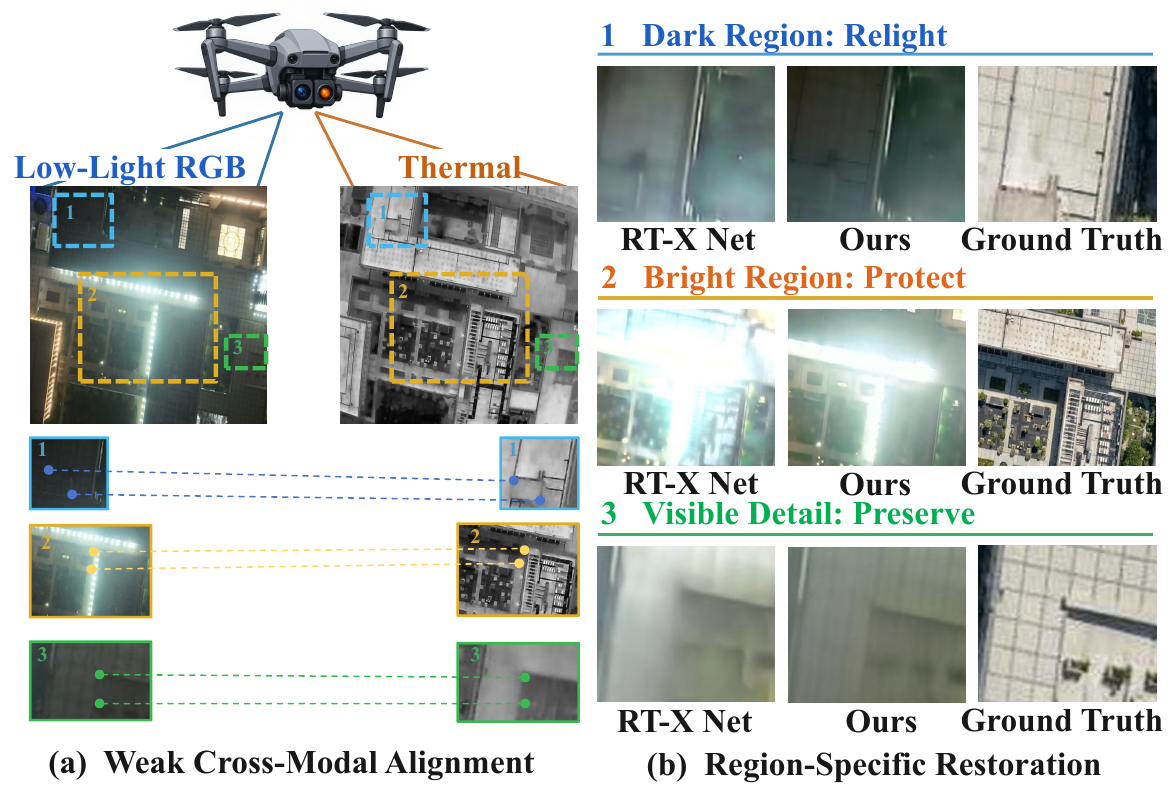}

    \caption[Motivation]{%
        Key challenges in UAV low-light image restoration with weakly aligned RGB and thermal images.
        (a) Cross-modal image pairs exhibit local spatial misalignment.
        Colored boxes indicate three corresponding regions, and dashed
        lines connect selected point correspondences in the enlarged
        patches.
        (b) The corresponding regions require different restoration
        strategies: relighting dark regions, protecting bright regions
        from over-enhancement, and preserving visible details.
        Results from RT-X Net and our method are shown alongside the
        ground truth.
        Matching colors and numbers associate the regions across
        the two panels.
    }
    \label{fig:motivation}
\end{figure}

Spatial correspondence alone does not determine the usefulness of thermal guidance. Thermal responses do not directly encode visible brightness or texture, and their relevance to visible-image restoration varies across the scene. Where visible details remain usable, large corrections may be unnecessary; where illumination is insufficient but structures remain discernible, luminance recovery is needed; where visible structures are weak, thermal cues may support restoration if their correspondence is reliable. Thermal guidance should also be attenuated where cross-modal responses are inconsistent. Fig.~\ref{fig:motivation} illustrates over-enhancement, loss of fine structure, and spatially uneven restoration, highlighting the need to accommodate different local conditions. The second challenge is therefore to adapt restoration to the remaining visible evidence, the degree of degradation, and the estimated reliability of thermal guidance.

Existing thermal-guided enhancement methods demonstrate the value of complementary thermal observations for restoring degraded visible images \citep{cao2023deep,wang2025thermal,jha2025rt}. For weakly aligned UAV imagery, exploiting this complementarity requires joint consideration of spatial correspondence, the remaining visible evidence, and local restoration needs. This motivates coupling correspondence and reliability estimation with region-adaptive restoration. Evaluating these requirements also calls for suitable data. Existing benchmarks support visible-only enhancement \citep{lolv2,u3d}, optical–thermal fusion and perception \citep{roadscene,m3fd}, or thermal-guided enhancement with normal-light references \citep{wang2025thermal}. Dedicated benchmarks for examining low-light UAV restoration in the presence of residual optical–thermal misalignment remain limited.

To address these challenges, we propose the Reliability-Aware Multi-Strategy Guidance Network (\textbf{RMGNet}) for thermal-guided restoration of low-light UAV imagery under weak cross-modal alignment. The framework jointly considers the spatial reliability of thermal guidance and the usability of the remaining visible evidence. The Reliability-Aware Guidance Calibration Module (\textbf{RAGCM}) uses visible and thermal structural cues to estimate local correspondence, evidence reliability, and spatial routing weights. Guided by these estimates, the Multi-Strategy Enhancement Routing Module (\textbf{MSERM}) combines detail preservation, highlight-protected relighting, and thermal-assisted compensation according to local restoration needs. The enhanced RGB image is produced in the coordinate system of the visible input. To support evaluation, we construct the UAV weakly aligned Optical-Thermal dataset (\textbf{UWOT}), comprising 5,000 low-light RGB, thermal, and normal-light reference triplets. UWOT combines UAV-acquired observations and DroneVehicle-derived samples, including both real nighttime and synthetically degraded low-light images.

The main contributions are summarized as follows:
\begin{itemize}
    \item We propose RMGNet, a thermal-guided framework for low-light UAV image enhancement that jointly considers uncertain cross-modal correspondence and spatially varying visible evidence to adapt thermal guidance to local restoration needs.

    \item We design RAGCM to perform structural-prior-guided coarse-to-fine alignment and estimate both thermal guidance reliability and visible evidence usability, supporting reliable cross-modal information transfer under residual misalignment.

    \item We introduce MSERM to adaptively combine detail preservation, highlight-protected relighting, and thermal compensation through spatial routing, addressing diverse local restoration requirements based on reliability and degradation cues.

    \item We construct UWOT with 5,000 low-light visible, thermal, and normal-light reference triplets. By combining real nighttime observations with synthetically degraded samples, UWOT supports UAV image enhancement research under residual optical-thermal misalignment.
\end{itemize}

\section{Related Work}
\label{sec:related_work}

In this section, we review related studies on visible-image enhancement, thermal-guided restoration, and cross-modal correspondence in UAV imagery. Particular attention is given to the observations available to each method, the restoration or interpretation objectives, and the correspondence and reference conditions provided by existing benchmarks.

\subsection{Low-Light Visible Image Enhancement}
\label{subsec:visible_llie}

Low-Light visible image enhancement addresses several interacting degradations, including insufficient illumination, noise, color distortion, and blur. Beyond enhancement of processed RGB images, \citet{chen2018sid}
investigate extreme low-light imaging directly in the RAW domain,
learning a mapping from short-exposure observations to
long-exposure references. Retinex-based approaches separate illumination adjustment from reflectance estimation \citep{Chen2018Retinex}, while Retinexformer incorporates illumination information into learned restoration \citep{Retinexformer}. SNR-aware enhancement adapts processing to local signal quality, accounting for spatial differences in noise and usable image content \citep{xu2022snr}. Other methods address complementary aspects: HVI-CIDNet separates color and intensity representations \citep{hvi}, DarkIR jointly addresses low illumination and blur \citep{darkir}, and MambaLLIE models global and local image dependencies \citep{MambaLLIE}. Zero-DCE formulates low-light enhancement as image-specific curve estimation and learns pixel-wise enhancement curves without reference images, providing an alternative to paired-supervision-based restoration \citep{guo2020zerodce}. These developments provide useful foundations for restoring degraded UAV visible imagery. 

UAV-oriented enhancement research further considers the acquisition and processing conditions of aerial imagery. \citet{u3d} introduce a benchmark and framework for ultra-high-resolution UAV low-light enhancement, addressing nonuniform illumination, limited paired data, and deployment constraints. Such work highlights the importance of evaluating enhancement under aerial imaging conditions. For image interpretation, the visibility of small objects and the continuity of fine structures are also relevant considerations. Visible-only methods, however, must estimate these structures from the degraded RGB observation and learned priors. When the observation contains little usable structural evidence, complementary sensing modalities offer an additional source of information. 

\subsection{Thermal-Guided Low-Light Enhancement}
\label{subsec:thermal_llie}

Thermal-guided enhancement uses complementary thermal observations to support visible-image restoration under insufficient illumination. TGLLE-Net extracts thermal edge and texture features at multiple receptive fields and applies spatially varying thermal-guided convolution to enhance degraded visible features \citep{cao2023deep}. RT-X Net uses cross-attention to exchange information between visible and thermal representations for nighttime enhancement \citep{jha2025rt}. \citet{wang2025thermal} develop a lightweight thermal-aware model that predicts normal-light images from low-light RGB and thermal inputs. VIIS instead formulates a joint information-enhancement and fusion task for severely degraded observations \citep{VIIS}. These approaches differ in both their information-transfer mechanisms and their output objectives, which should be considered when relating them to reference-based RGB restoration.

Multimodal restoration has also been investigated specifically for UAV observations. \citet{wpid} combine visible–thermal fusion with weakly paired training in HDCGAN+ and evaluate structural consistency using segmentation outputs. In optics-guided thermal UAV super-resolution, GDNet adapts optical guidance to adverse conditions, including low illumination and fog \citep{Zhao2025GDNet}. Although its reconstruction target is thermal imagery, this study demonstrates the relevance of guidance quality to multimodal aerial restoration. Together, these works motivate distinguishing the quality of the guiding observation, its spatial correspondence with the target modality, and the desired output representation. The present work focuses on visible-image restoration with residual optical–thermal misalignment and locally varying evidence reliability.

\subsection{Weakly Aligned RGB-T Fusion and UAV Benchmarks}
\label{subsec:alignment_benchmarks}

Cross-modal correspondence has been studied in both UAV image registration and multimodal perception. \citet{Meng2022TWMM} address the registration of thermal infrared and visible images acquired by UAV dual-camera systems, accounting for substantial differences in their image responses. For UAV object detection, \citet{chen2024oafa} estimate cross-modal feature offsets to accommodate displaced targets, while DroneVehicle supports RGB–infrared vehicle detection and uncertainty-aware fusion \citep{dronevehicle}. These studies address related correspondence problems under different objectives. Image registration seeks spatial agreement, whereas detection-oriented alignment supports object recognition and localization. Thermal-guided RGB restoration additionally requires correspondence estimates to support visible appearance reconstruction and local detail preservation.

Existing benchmarks provide different combinations of observations and reference conditions. TLIE contains low-light visible images, normal-light references, and thermal observations for thermal-guided enhancement \citep{wang2025thermal}. V-TIEE provides co-located optical–thermal observations under nighttime conditions \citep{jha2025rt}, while U3D focuses on ultra-high-resolution UAV visible-image enhancement \citep{u3d}. WPID includes daytime visible, nighttime visible, and thermal imagery, with weak pairing primarily referring to the correspondence between daytime and nighttime visible observations \citep{wpid}. These settings distinguish two correspondence requirements: the relation between low-light inputs and their restoration references, and the relation between visible and thermal inputs. UWOT complements these resources by providing 5,000 UAV-oriented triplets for studying thermal-guided visible-image restoration with retained cross-modal spatial discrepancies.

\section{Methodology}
In this section, we introduce the proposed Reliability-Aware Multi-Strategy Guidance Network (RMGNet) for thermal-guided low-light UAV image enhancement. We first present the overall framework, followed by pre-cleaning and structural prior extraction. We then describe the Reliability-Aware Guidance Calibration Module (RAGCM) and the Multi-Strategy Enhancement Routing Module (MSERM), which calibrate thermal guidance and adaptively combine complementary restoration strategies, respectively. Finally, we introduce the loss function used to train the network.

\begin{figure*}[pos=t]
    \centering
    \includegraphics[width=\textwidth]{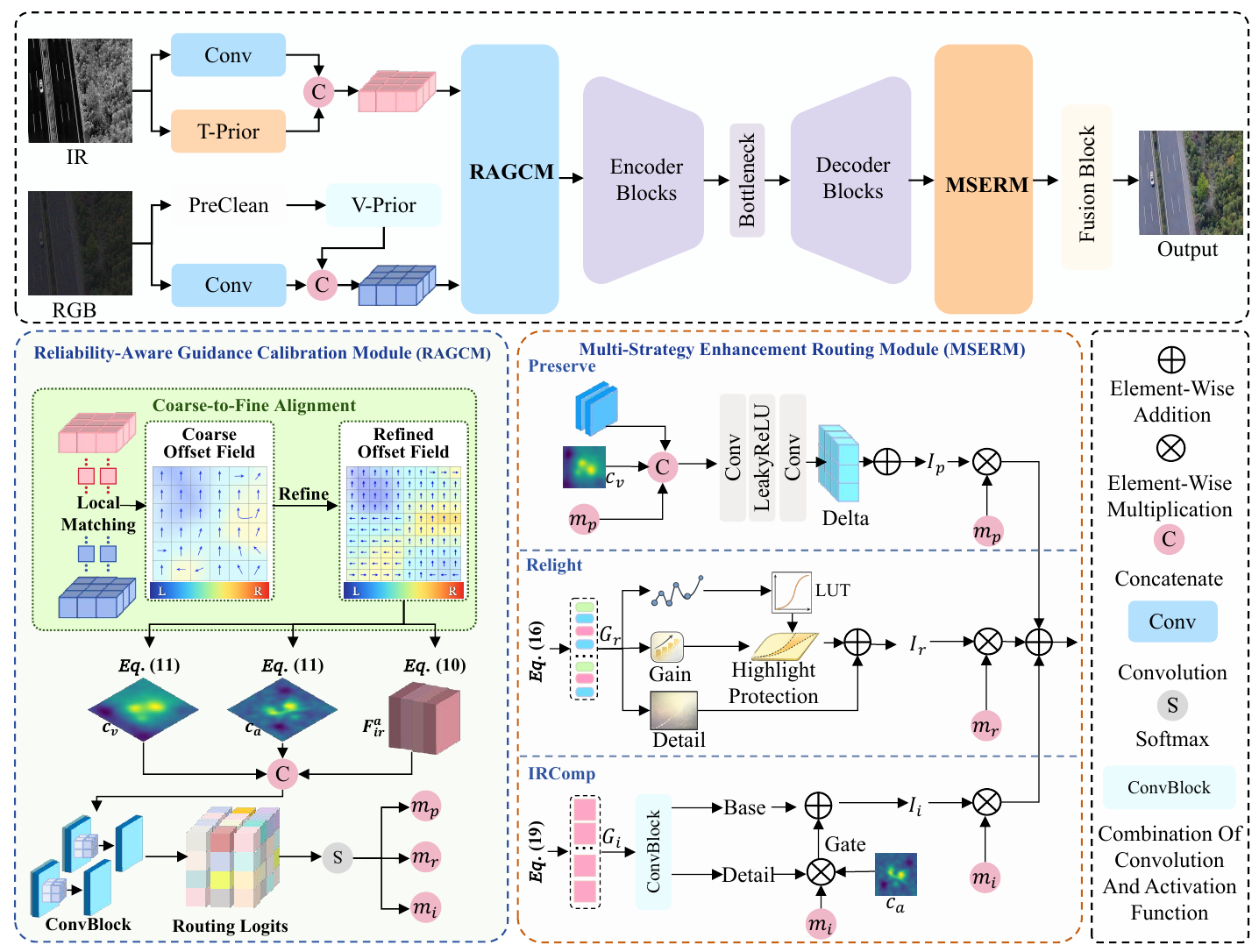}
    \caption{Overall pipeline of the proposed RMGNet for thermal-guided low-light UAV image enhancement. The front end extracts shallow features, degradation cues, and visible and thermal structural priors. The Reliability-Aware Guidance Calibration Module (RAGCM) performs coarse-to-fine alignment and predicts alignment confidence, visible confidence, and spatial routing weights. An encoder–decoder backbone produces a preliminary restoration, which the Multi-Strategy Enhancement Routing Module (MSERM) refines through detail preservation, highlight-protected relighting, and thermal compensation. The three candidates are combined using the predicted routing weights to obtain the final enhanced image. The lower panels detail the calibration and candidate-generation processes.}
    \label{fig:RAGCM}
\end{figure*}

\subsection{Overall Framework}
\label{subsec:overall_framework}

Given a low-light visible image $I_{\mathrm{rgb}}$ and its paired thermal observation $I_{\mathrm{ir}}$, RMGNet estimates an enhanced visible image $\hat{I} = f_{\theta}(I_{\mathrm{rgb}}, I_{\mathrm{ir}})$. The visible image defines the output grid, whereas thermal features provide auxiliary structural information under imperfect cross-modal correspondence. The model therefore combines local correspondence estimation with adaptive control of thermal guidance. Data acquisition, reference construction, and synthetic degradation are described in Section~4.

As illustrated in Fig. ~\ref{fig:RAGCM}, RMGNet first extracts shallow features, degradation cues, and modality-specific structural priors. A pre-cleaning stage prepares the visible input for structural cue extraction. RAGCM then estimates local visible–thermal correspondence and predicts guidance confidence and spatial routing weights. The aligned thermal features are modulated before initial fusion with visible features, and an encoder–decoder backbone produces a preliminary reconstruction. MSERM subsequently generates three enhancement candidates, allowing bounded refinement, luminance adjustment, and additional thermal-guided compensation to contribute differently across the image.

RMGNet is optimized end-to-end using the restoration references and the training objective described in Section~3.5. During inference, a low-light visible image and its thermal counterpart are supplied to the trained network to produce an enhanced visible image in the coordinate system of the visible input. The confidence and routing maps are predicted directly from the inputs, allowing the contribution of thermal guidance and the combination of enhancement strategies to vary across image regions.

\subsection{Pre-Cleaning and Structural Prior Extraction}

Low-light degradation can obscure narrow markings and object boundaries in UAV imagery, while noise may introduce spurious gradient responses. We therefore apply residual pre-cleaning before extracting visible structural cues for cross-modal matching. This stage provides a regularized input for correspondence estimation, with the correction magnitude controlled by the residual scale in Eq. (1). 

Given a low-light visible image $I_{rgb}$, the pre-cleaning module extracts a feature $F_c$ and predicts a bounded residual to obtain a mildly regularized image $I_c$:
\begin{equation}
\begin{gathered}
F_c = \mathcal{B}(\mathcal{S}(I_{rgb})),\\
I_c = \mathrm{clip}\big(I_{rgb}+\tanh(\phi_c(F_c))\cdot s_c,\,0,\,1\big),
\end{gathered}
\end{equation}
where $\mathcal{S}(\cdot)$ and $\mathcal{B}(\cdot)$ denote the stem and body, $\phi_c(\cdot)$ is the residual head, and $s_c$ is a learnable residual scale. From the same representation, we estimate a noise cue $N$ and a darkness cue $D$. The darkness cue $D$ describes local brightness deficiency in the pre-cleaned visible image, while $N$ provides a learned noise-related indicator. Together, these cues condition subsequent correspondence refinement and enhancement on the local state of the visible observation. They are defined as follows:
\begin{equation}
\begin{gathered}
N=\sigma\big(\phi_n(F_c)+|I_{rgb}-I_c|_{\mathrm{mean}}\big), \\
D=1-\mathrm{Luma}(I_c).
\end{gathered}
\end{equation}
The pre-cleaned image $I_c$ stabilizes structure extraction, while $N$ and $D$ retain local degradation information.

\begin{figure}[pos=t]
    \centering
    \includegraphics[width=\linewidth]
    {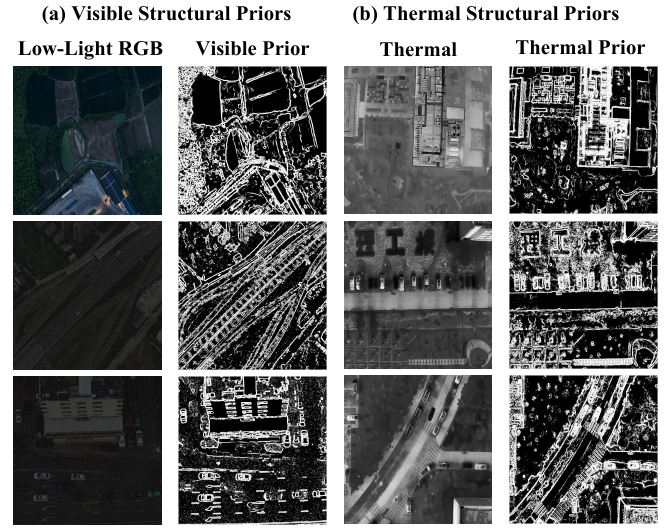}
    \caption{
        Structural priors for cross-modal correspondence estimation. (a) Low-light RGB images and their corresponding visible structural-prior maps. (b) Thermal images and their corresponding thermal structural-prior maps. In each subfigure, the input images and prior maps are shown in the left and right columns, respectively. The prior maps highlight edges and contours within each modality.
    }
    \label{fig:structural_priors}
\end{figure}

We then construct visible and thermal structural priors from gradient and Laplacian responses:
\begin{equation}
\begin{gathered}
G_c = \mathrm{Sobel}(I_c), \quad L_c = \mathrm{Lap}(I_c),\\
(P_v, F_v^p) = \Phi_v(I_c, G_c, L_c, F_c, N, D),\\
G_t = \mathrm{Sobel}(I_{ir}), \quad L_t = \mathrm{Lap}(I_{ir}),\\
(P_t, F_t^p) = \Phi_t(I_{ir}, G_t, L_t).
\end{gathered}
\end{equation}
Here, $P_v$ and $P_t$ are map-level visible and thermal structural priors, while $F_v^p$ and $F_t^p$ denote feature-level priors. As shown in Fig.~\ref{fig:structural_priors}, the structural-prior maps highlight modality-specific edges and contours, providing structural cues for subsequent cross-modal correspondence estimation. These priors, together with the shallow visible and thermal features and the degradation cues $(D,N)$, are fed into RAGCM for guidance calibration under weak optical-thermal alignment. 

\subsection{Reliability-Aware Guidance Calibration Module}

Thermal guidance can become misleading when its structures are spatially displaced from the visible observation or provide weak correspondence to visible scene content. RAGCM therefore estimates local thermal-to-visible feature correspondence and the reliability of the available evidence. Here, guidance calibration refers to feature correspondence estimation and adaptive weighting of thermal guidance. The module takes as inputs the shallow visible and thermal features $F_r^0$ and $F_i^0$, the structural-prior features $F_v^p$ and $F_t^p$, the prior maps $P_v$ and $P_t$, and the degradation cues $D$ and $N$. It outputs an aligned thermal feature $F_{\mathrm{ir}}^a$, an alignment confidence map $c_a$, a visible confidence map $c_v$, and a spatial routing map $m$ for subsequent restoration.

Coarse correspondence is estimated by comparing structural features within a local displacement window. To construct comparable matching representations, the visible and thermal prior features are first projected into a shared normalized correlation space:
\begin{equation}
\bar{Q}_v = \operatorname{Norm}(\psi_v(F_v^p)), \qquad
\bar{Q}_t = \operatorname{Norm}(\psi_t(F_t^p)).
\end{equation}
For each spatial location, $\bar{Q}_t$ is shifted within a local search window 
\[
\Omega =
\left\{
(\Delta x,\Delta y)\in\mathbb{Z}^{2}
\,\middle|\,
|\Delta x|\leq r,\ |\Delta y|\leq r
\right\}.
\]
We set the search radius to $r=2$, resulting in a $5\times5$ window with $(2r+1)^2=25$ candidate displacements at each spatial location. For each displacement $\delta\in\Omega$, the correlation response and matching probability are computed as
\begin{equation}
\begin{gathered}
C_{\delta}(x) = \operatorname{mean}_{c}
\left(\bar{Q}_v(x)\odot \bar{Q}_t(x+\delta)\right),\\
p_{\delta}(x) = \frac{\exp(C_{\delta}(x))}
{\sum_{\delta' \in \Omega}\exp(C_{\delta'}(x))}.  
\end{gathered}
\end{equation}
The matching distribution represents competing local correspondence hypotheses. The coarse displacement is computed as the expectation over these hypotheses:
\begin{equation}
o_c(x)=\sum_{\delta \in \Omega} p_{\delta}(x)\,\delta .
\end{equation}
Peak and entropy statistics, denoted by $s_p$ and $s_e$, summarize the concentration and ambiguity of the matching distribution. These statistics provide additional matching cues for subsequent displacement refinement.

For fine alignment, the thermal feature $F_i^0$, structural-prior feature $F_t^p$, and prior map $P_t$ are first warped toward the visible-image grid using the coarse displacement $o_c$:
\begin{equation}
\tilde{F}_i=\mathcal{W}(F_i^0,o_c),\quad
\tilde{F}_t^p=\mathcal{W}(F_t^p,o_c),\quad
\tilde{P}_t=\mathcal{W}(P_t,o_c),
\end{equation}
where $\mathcal{W}(\cdot,\cdot)$ denotes differentiable warping. Local shifts used to construct the correlation volume employ replicated boundary padding, while differentiable warping uses bilinear grid sampling with border padding. The structural residual is defined as
\begin{equation}
\Delta_s = |P_v-\tilde{P}_t|.
\end{equation}
Residual feature differences and structural discrepancies are combined with visible degradation cues to refine the remaining local mismatch. This conditions the refinement on both cross-modal agreement and the quality of the visible evidence. The refinement representation is constructed as follows:
\begin{equation}
\mathcal{R}=
\left[
F_r^0,\tilde{F}_i, |F_r^0-\tilde{F}_i|, F_v^p,
\tilde{F}_t^p, D,N,\Delta_s,s_p,s_e
\right].
\end{equation}
The offset refinement, final displacement, and aligned thermal feature are then computed as
\begin{equation}
\begin{gathered}
\Delta o
=
d_{\max}\widetilde{s}_o
\tanh\left(\phi_{\mathrm{ref}}(\mathcal{R})\right),\\
o=o_c+\Delta o,\\
F_{ir}^{a}=\mathcal{W}(F_i^0,o),
\end{gathered}
\end{equation}
where $\phi_{\mathrm{ref}}(\cdot)$ is a lightweight offset-refinement network, $d_{\max}=4$ is the nominal displacement scale at the input feature resolution, and $\widetilde{s}_o=\operatorname{clip}(s_o,0.1,10)$ is a bounded learnable scale with $s_o$ initialized to 1.

RAGCM further predicts $c_a$ and $c_v$ to measure the local reliability of aligned thermal guidance and visible evidence. Here, $c_a$ represents the pixel-wise reliability of the aligned thermal feature, whereas $c_v$ measures the usability of the visible information under low-light degradation. After applying the refined displacement $o$, the alignment confidence is estimated from the resulting aligned representation $R_a$, while the visible confidence is estimated from the visible feature, visible prior feature, and degradation cues:
\begin{equation}
c_a=\sigma\!\left(\phi_a(R_a)\right),\qquad
c_v=\sigma\!\left(\phi_v([F_r^0,F_v^p,D,N])\right),
\end{equation}
where $R_a$ denotes the final cross-modal representation obtained after applying the refined displacement $o$. RAGCM also predicts spatial weights for the three enhancement candidates. These weights allow the relative contribution of preservation, relighting, and thermal-guided compensation to vary across image regions. The routing map is predicted as
\begin{equation}
m=\operatorname{softmax}(z/\tau)=[m_p,m_r,m_i],
\end{equation}
where $z$ denotes the routing logits and $\tau$ is a learnable temperature. Here, $m_p$, $m_r$, and $m_i$ provide the spatial weights for preservation, relighting, and thermal compensation in MSERM, respectively. These weights condition candidate generation and spatially adaptive fusion in MSERM.

\subsection{Multi-Strategy Enhancement Routing Module}

To account for uncertain optical-thermal correspondence during restoration, the contribution of aligned thermal features is adjusted before initial feature fusion. Specifically, the aligned thermal feature $F_{\mathrm{ir}}^a$ is modulated by the thermal-compensation routing weight $m_i$ and the alignment confidence $c_a$:
\begin{equation}
\begin{aligned}
g_{\mathrm{inj}}
&= (0.05 + 0.95m_i)
   \odot (0.25 + 0.75c_a), \\
F_{\mathrm{ir}}^g
&= F_{\mathrm{ir}}^a \odot g_{\mathrm{inj}}.
\end{aligned}
\label{eq:thermal_gate}
\end{equation}

The gate reduces the contribution of this thermal feature pathway when the predicted alignment confidence or thermal-compensation weight is low, while retaining a positive gate floor. The gated thermal feature is fused with the shallow visible feature to obtain $F^0$. The encoder-decoder backbone then produces a preliminary reconstruction $I_{\mathrm{pre}}$ and a shared decoder feature $F_d$:
\begin{equation}
\begin{aligned}
F^{0}
&= \phi_{\mathrm{fuse}}
   \bigl([F_r^{0}, F_{\mathrm{ir}}^{g}]\bigr), \\
(F_d, I_{\mathrm{pre}})
&= \operatorname{Backbone}
   \bigl(F^{0}, F_{\mathrm{ir}}^{a}, m_i, c_a\bigr).
\end{aligned}
\label{eq:guided_backbone}
\end{equation}

The shared multimodal representations are used to generate three candidates for bounded refinement, luminance adjustment, and additional thermal-guided compensation. These operations address spatially varying restoration needs, and their local contributions are controlled by the routing map predicted by RAGCM.

Some regions of the preliminary reconstruction may require only limited correction. The preservation candidate therefore applies a bounded residual update to the preliminary multimodal result $I_{\mathrm{pre}}$. The update is conditioned on the decoder feature $F_d$, preservation weight $m_p$, and visible confidence $c_v$:
\begin{equation}
\begin{gathered}
\Delta_p=\gamma_p \tanh\!\bigl(\phi_p([F_d,m_p,c_v])\bigr),\\
I_p=\operatorname{clip}\!\left(I_{pre}+\Delta_p\odot(\alpha_p+\beta_p m_p),\,0,1\right),
\end{gathered}
\end{equation}
where $\gamma_p$ is a learnable residual scale, and $\alpha_p = 0.2$ and $\beta_p = 0.8$ are fixed coefficients controlling the residual modulation strength. The initialization and bounds of $\gamma_p$ are specified in Section~5.1.

Spatially varying illumination can create different luminance-adjustment needs within a UAV image. The relighting candidate combines adaptive luminance adjustment with soft highlight protection. Its guidance representation is constructed from the decoder feature $F_d$, shallow fused feature $F^0$, relighting weight $m_r$, visible confidence $c_v$, and darkness cue $D$:
\begin{equation}
G_r=\phi_r([F_d,F^0,m_r,c_v,D]).
\end{equation}
We define a luminance-adjustment operator $\mathcal{T}(I;G)$ for an input image $I$ and guidance representation $G$. The operator combines a monotonic LUT with 33 knots, a spatial gain, soft highlight protection, and a bounded RGB detail residual. Let $y = \mathrm{Luma}(I)$. The adjusted luminance and RGB output are computed as follows:
\begin{equation}
\begin{aligned}
y_{\mathrm{loc}}
&= \operatorname{clip}\!\Bigl(
   \operatorname{LUT}_{33}(y;G) \\
&\qquad {}\times
   \exp\!\bigl(\gamma_g\tanh(\phi_g(G))\bigr),
   0,1\Bigr), \\
w_h
&= \sigma\!\bigl(14(y-0.72)\bigr), \\
y_{\mathrm{out}}
&= (1-w_h)\odot y_{\mathrm{loc}} + w_h\odot y, \\
r_y
&= \operatorname{clip}\!\left(
   \frac{y_{\mathrm{out}}}{y+\epsilon},0,16
   \right), \\
\mathcal{T}(I;G)
&= \operatorname{clip}\!\Bigl(
   I\odot r_y \\
&\qquad {}+
   \gamma_d\tanh(\phi_d(G)),0,1\Bigr).
\end{aligned}
\label{eq:luminance_adjustment}
\end{equation}

Positive LUT increments are generated using softplus, followed by cumulative summation and normalization, ensuring a monotonic LUT mapping. The spatial gain allows local luminance adjustment. Highlight protection limits luminance changes in bright regions, while the bounded luminance ratio controls amplification in very dark pixels. The relighting candidate is obtained as
\begin{equation}
I_r = \mathcal{T}(I_{\mathrm{pre}};G_r).
\label{eq:relighting_candidate}
\end{equation}

The thermal-compensation candidate provides an additional adjustment conditioned on aligned thermal features and visible degradation cues. It reuses the gated thermal feature $F_{\mathrm{ir}}^g$ defined in Eq.~\eqref{eq:thermal_gate}, which is also used for initial feature fusion.
The thermal-guided representation is constructed as follows:
\begin{equation}
G_i
=
\phi_i\left(
\left[
F_d,
F_{\mathrm{ir}}^g,
m_i,
c_a,
D,
N
\right]
\right),
\label{eq:thermal_guided_representation}
\end{equation}
where $N$ denotes the noise cue. A base result $I_b$ adopts the same monotonic luminance design as the relighting candidate, and a confidence-gated compensation residual is added:
\begin{equation}
\begin{aligned}
\Delta_i &= \gamma_i \tanh\!\left(\varphi_{di}(G_i)\right), \\
I_i &= \operatorname{clip}
\left(I_b + \Delta_i \odot g_i,\, 0,\, 1\right), \\
g_i &= (\alpha_c + \beta_c m_i)
       \odot (\eta_c + \kappa_c c_a).
\end{aligned}
\label{eq:thermal_compensation}
\end{equation}
Here, $\alpha_c=\eta_c=0.2$ and $\beta_c=\kappa_c=0.8$ are fixed coefficients that remain unchanged during both training and inference. The thermal-compensation routing weight $m_i$ and the alignment confidence $c_a$ are predicted by the network. Since $m_i$ and $c_a$ lie in $[0,1]$, each affine gating factor is bounded between $0.2$ and $1$, and their element-wise product satisfies $0.04 \leq g_i \leq 1$. This gate attenuates the compensation residual in regions with low thermal-compensation routing weights or low alignment confidence, while retaining a nonzero minimum gate to avoid completely suppressing the compensation pathway.

The three candidate images are combined on the visible-image grid using their spatial routing weights: 
\begin{equation}
\hat{I}=\operatorname{clip}\!\left(m_p\odot I_p+m_r\odot I_r+m_i\odot I_i,0 , 1\right).
\end{equation}
Here, $m_p$, $m_r$, and $m_i$ are the learned, softmax-normalized routing weights for preservation, relighting, and thermal compensation, respectively. They sum to one at each spatial location, allowing the contribution of each enhancement operation to vary across the image. The final weighted summation introduces no additional learnable parameters.

\subsection{Loss Function}
We train the whole network with a unified objective consisting of four groups of constraints, including an L1-family loss, a perceptual loss, an edge-aware smoothness regularization, and a heuristic correlation constraint:
\begin{equation}
L = 
\mathcal{L}_{L_1}
+
\mathcal{L}_{perc}
+
\mathcal{L}_{smooth}
+
\mathcal{L}_{heur}.
\end{equation}
Here, \(\mathcal{L}_{L_1}\) denotes the L1-family loss applied to image restoration, candidate supervision, structural consistency, and alignment-related variables. $\mathcal{L}_{per}$ improves perceptual similarity. $\mathcal{L}_{smooth}$ regularizes the spatial continuity of the confidence and routing maps. $\mathcal{L}_{heur}$ encourages reliable local correlation matching. Detailed formulations and weighting coefficients for each loss term are provided in the supplementary material.

\section{Dataset}
In this section, we introduce the UAV weakly aligned Optical-Thermal (UWOT) dataset for thermal-guided low-light image enhancement. UWOT comprises triplets of low-light visible images, normal-light references, and thermal images, primarily collected using a DJI Mavic 3T UAV and supplemented with samples derived from DroneVehicle. 

\begin{figure*}[pos=t]
    \centering
    \includegraphics[width=\linewidth]{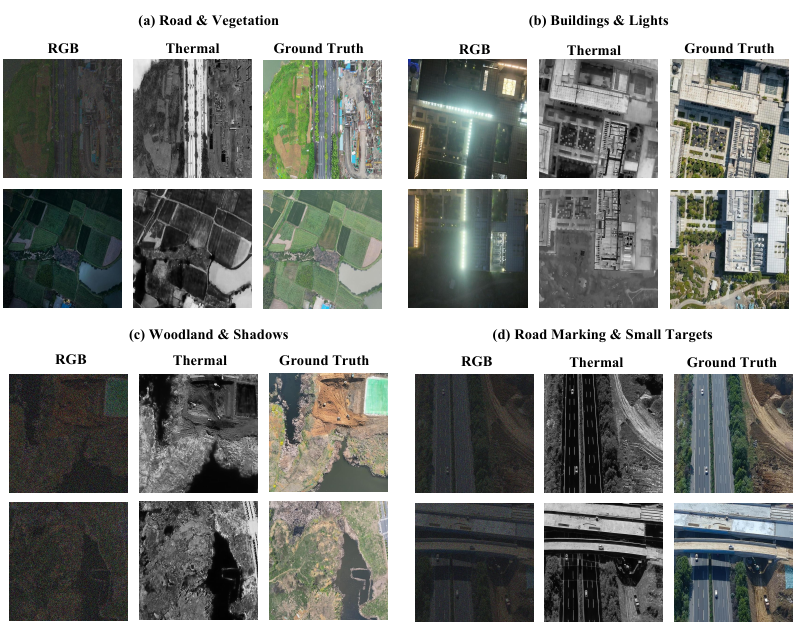}
    \caption{
    Representative examples from the UAV weakly aligned Optical-Thermal
(UWOT) dataset. The samples are organized into four scene groups:
(a) roads and vegetation, (b) buildings and lights, (c) woodland and shadows, and (d) road markings and small targets. Each panel contains two triplets, with each row showing, from left to right, a low-light RGB input, its paired thermal observation, and a normal-light RGB reference. The optical-thermal pairs retain residual spatial misalignment.
}
    \label{fig:uwot_samples}
\end{figure*}

\begin{figure}[pos=t]
    \centering
    \includegraphics[width=\columnwidth]{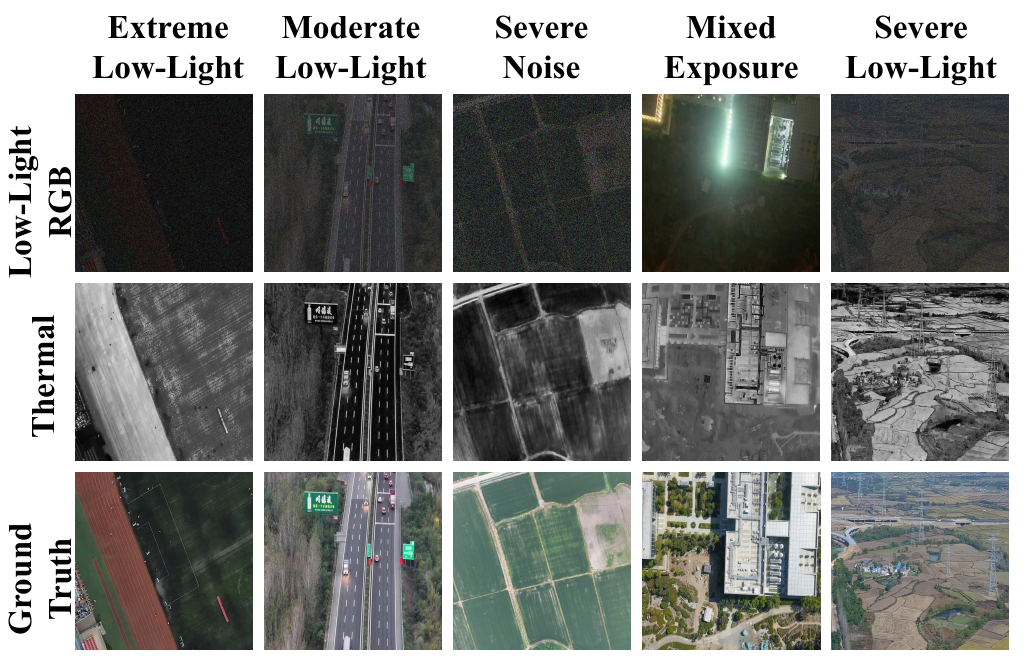}
    \caption{
    Representative low-light degradation scenarios in the UWOT dataset.
    The presented samples cover five typical degradation conditions: extreme low-light, moderate low-light, severe noise degradation, mixed exposure, and severe low-light. Each sample consists of a low-light RGB image, its corresponding thermal observation, and a normal-light RGB reference. These examples demonstrate the diversity of illumination conditions and noise degradation in the dataset, while highlighting the complementary structural information provided by thermal imaging.
    }
    \label{fig:uwot_degradation}
\end{figure}

\begin{table*}[t]
\centering

\caption{
Comparison of UWOT with representative visible low-light and optical-thermal datasets. “Sample Unit” specifies whether the dataset size is counted in individual images, image pairs, or image triplets. ``LL RGB'' and ``Normal RGB'' denote low-light and normal-light images, respectively. ``Weak Align.'' indicates that residual misalignment between visible and thermal observations is retained. ``Public'' indicates whether the dataset is publicly downloadable. $^{*}$The weak pairing in WPID primarily concerns daytime and nighttime visible images. Its nighttime optical-thermal pairs undergo geometric registration during preprocessing; therefore, WPID is not marked as a benchmark explicitly targeting weak RGB--T alignment.
}
\label{tab:dataset_compare}

\small
\setlength{\tabcolsep}{6pt}
\renewcommand{\arraystretch}{1.08}

\begin{tabular}{lcccccccc}
\toprule

\rowcolor{headerblue}
Dataset
& Size
& Sample Unit
& LL RGB
& Thermal
& Normal RGB
& UAV
& Weak Align.
& Public \\
\midrule

LOLv2-real \citep{lolv2}
& 789
& Image Pair
& \checkmark
& --
& \checkmark
& --
& --
& \checkmark \\

LOLv2-syn \citep{lolv2}
& 1,000
& Image Pair
& \checkmark
& --
& \checkmark
& --
& --
& \checkmark \\

U3D \citep{u3d}
& 2,000
& Single Image 
& \checkmark
& --
& --
& \checkmark
& --
& \checkmark \\

RoadScene \citep{roadscene}
& 221
& Image Pair
& --
& \checkmark
& --
& --
& --
& \checkmark \\

TLIE \citep{wang2025thermal}
& 1,113
& Image Triplet
& \checkmark
& \checkmark
& \checkmark
& --
& --
& \checkmark \\

M3FD \citep{m3fd}
& 4,200
& Image Pair
& --
& \checkmark
& --
& --
& --
& \checkmark \\

WPID \citep{wpid}
& 517
& Image Triplet
& \checkmark
& \checkmark
& \checkmark
& \checkmark
& --$^{*}$
& -- \\

\rowcolor{lastgray}
\textbf{UWOT}
& \textbf{5,000}
& Image Triplet
& \checkmark
& \checkmark
& \checkmark
& \checkmark
& \checkmark
& \checkmark \\

\bottomrule
\end{tabular}

\end{table*}

\subsection{UAV Acquisition and Observation Conditions}

For real-world samples, daytime visible images are captured as normal-light references, while nighttime visible and thermal infrared images are collected at the same locations. The flight position, altitude, viewing angle, and heading are kept as consistent as possible. This protocol reflects practical UAV nighttime imaging, where low-light degradation and optical-thermal misalignment often coexist.

\subsection{Controlled Synthesis of Low-Light Observations}

The low-light visible images include both real nighttime captures and synthetically degraded samples. For the synthetic subset, each normal-light visible image is first converted from sRGB to an approximate linear-RGB domain using the piecewise inverse sRGB transfer
function \citep{anderson1996srgb}. For each image, we independently sample one darkening factor $\alpha$, one read-noise level $\sigma_r$, and one signal-dependent noise coefficient $\beta$. These scalar parameters are shared across all pixels and RGB channels within the image.

Given the linear normal-light image
$J_n=\mathcal{T}_{\mathrm{sRGB}}^{-1}(I_n)$, we obtain the darkened signal as
\begin{equation}
    J_d = \alpha J_n,
    \qquad
    \alpha \sim \mathcal{U}(0.05,0.08).
    \label{eq:syn_darkening}
\end{equation}
Independent additive read noise is sampled for every pixel position $p$ and color channel $c$ as
\begin{equation}
    n_r(p,c)
    \sim
    \mathcal{N}
    \left(0,\left(\frac{\sigma_r}{255}\right)^2\right),
    \qquad
    \sigma_r \sim \mathcal{U}(2.5,5.0).
    \label{eq:syn_read_noise}
\end{equation}
Signal-dependent noise is modeled using a heteroscedastic Gaussian approximation to shot noise:
\begin{equation}
    n_s(p,c)
    \sim
    \mathcal{N}
    \left(
        0,
        \beta^2 J_d(p,c)
    \right),
    \qquad
    \beta \sim \mathcal{U}(0.03,0.09).
    \label{eq:syn_shot_noise}
\end{equation}
The degraded linear image is then computed as
\begin{equation}
    \widetilde{J}_l=\operatorname{clip}
    \left( J_d+n_r+n_s,\,0,\,1\right).
    \label{eq:syn_degradation}
\end{equation}
It is subsequently mapped back to the sRGB domain using the standard forward sRGB transfer function, clipped to $[0,1]$, converted to an 8-bit representation, and saved to disk.

The synthetic degradation is performed offline. Each normal-light image produces one fixed low-light realization, and the degradation parameters are sampled once during dataset construction rather than resampled during training. Noise samples are independent across spatial locations and RGB channels. 

\subsection{Construction of DroneVehicle-Derived Triplets}

To increase scene diversity, we additionally construct triplets from the public DroneVehicle dataset. For each original paired optical-thermal observation, the daytime visible image is treated as the normal-light reference $I_n$, and its paired infrared image is retained unchanged as the thermal guidance image $I_{\mathrm{ir}}$. The low-light visible image $I_l$ is generated offline from $I_n$ using the synthetic degradation process described above. Each DroneVehicle-derived sample is thus organized as $(I_l, I_{\mathrm{ir}}, I_n)$. The object-detection annotations provided by DroneVehicle are not used in either network training or evaluation.

Since $I_l$ is directly synthesized from $I_n$, the low-light and normal-light RGB images are pixel-wise aligned. By contrast, the thermal image preserves the original optical-thermal correspondence and its residual cross-modal displacement. We only perform cropping, resizing, and format conversion for dataset unification and do not apply dense pixel-level optical-thermal registration. Consequently, these triplets provide accurately aligned low-light/normal-light RGB supervision while retaining weak optical-thermal alignment.

\subsection{Weak Visible-Thermal Correspondence}
\label{sec:weak_correspondence}

Visible and thermal images in UWOT exhibit weak spatial alignment due to parallax, sensor offsets, differences in resolution and field of view, and platform motion. These factors introduce residual spatial discrepancies even when the two modalities observe the same scene. During dataset preparation, we perform necessary cropping, triplet organization, and format unification without applying dense pixel-level optical-thermal registration. This preserves scene- and object-level correspondence while retaining local misalignment encountered in UAV imaging.

The optical-thermal correspondence should be distinguished from the correspondence between low-light RGB images and their normal-light references.
For synthetically degraded samples, the low-light image is generated directly from its normal-light RGB reference without geometric transformation, preserving pixel-wise RGB correspondence. The paired thermal image is retained unchanged and therefore preserves the original cross-modal displacement. For real captures, daytime references and nighttime observations are acquired at the same locations with flight configurations kept as consistent as possible, as described in the acquisition protocol.

Retaining residual optical-thermal misalignment makes UWOT suitable for evaluating thermal guidance under uncertain spatial correspondence. In severely degraded regions, visible structures may provide unreliable matching cues, while thermal structures may be informative but spatially displaced. The benchmark therefore examines how effectively enhancement methods recover visible content while controlling the influence of imperfectly aligned thermal guidance.

\subsection{Dataset Statistics and Benchmark Characteristics}
\label{sec:dataset_statistics}

UWOT contains 5,000 optical-thermal triplets, comprising 15,000 images in total. Each triplet consists of a low-light RGB image, a thermal image, and a normal-light RGB reference. The dataset is divided into 4,000 training triplets, 500 validation triplets, and 500 testing triplets. Low-light RGB and thermal images serve as the paired inputs, while normal-light RGB images provide references for restoration training and evaluation.

The dataset combines UAV-acquired observations with additional triplets derived from DroneVehicle, and includes both real nighttime images and synthetically degraded low-light images. For the synthetic subset, degradation is performed offline, producing one fixed low-light realization for each normal-light reference. Representative triplets are shown in Fig.~\ref{fig:uwot_samples}, illustrating diverse scene structures, illumination conditions, and residual optical-thermal displacement.

In addition to scene diversity, UWOT encompasses a range of low-light degradation conditions. As illustrated in Fig.~\ref{fig:uwot_degradation}, representative samples exhibit different degradation characteristics, including extreme low-light, moderate low-light, severe noise degradation, mixed exposure, and severe low-light. These examples highlight variations in illumination deficiency, noise contamination, and local exposure imbalance across the dataset. The corresponding thermal observations provide complementary structural information that may be difficult to recover from degraded visible images alone.

As summarized in Table~\ref{tab:dataset_compare}, UWOT brings together UAV imaging, low-light RGB observations, normal-light references, and weakly aligned thermal guidance within a single benchmark. Unlike visible-only LLIE datasets, it supports the evaluation of complementary thermal information. Unlike RGB-T datasets primarily designed for detection, segmentation, or fusion, it targets the restoration of visible image appearance under low illumination. Its evaluation setting combines visible degradation with cross-modal spatial uncertainty, enabling assessment of restoration quality when thermal guidance is available but imperfectly aligned.

\section{Experiments}
In this section, we evaluate the proposed RMGNet on the UWOT and a DroneVehicle-derived benchmark. We first introduce the experimental settings, baseline construction, and comparison protocol, followed by quantitative and qualitative comparisons with representative single-modal and multimodal methods. We then analyze the learned confidence and routing maps and evaluate robustness to cross-modal misalignment. Finally, we conduct ablation studies to examine the contributions of the main modules, enhancement candidates, and loss components.

\subsection{Experimental Settings}
\label{sec:experimental_settings}

\begin{figure*}[pos=t]
    \centering
    \includegraphics[width=\textwidth]{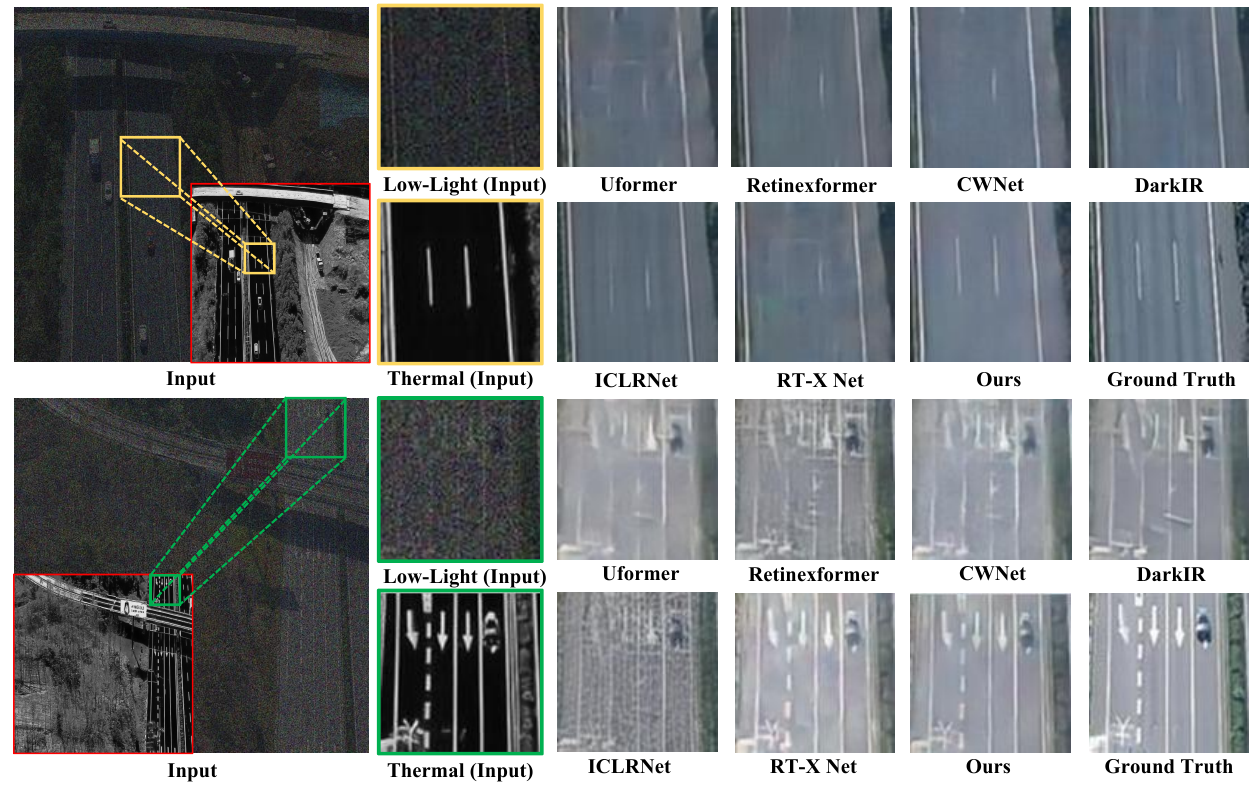}
    \caption{Qualitative comparison on UWOT test samples. Red-bordered insets show thermal inputs; yellow and green boxes mark the regions enlarged on the right.}
    \label{fig:qualitative_comparison}
\end{figure*}

\begin{table*}[t]
\centering

\caption{
Quantitative comparison with representative single-modal and multimodal low-light image enhancement methods on UWOT. The best and second-best results are shown in bold and underlined, respectively. "-M" denotes a thermal-input variant constructed in this work. 
}
\label{tab:quantitative_comparison}

\small
\setlength{\tabcolsep}{5pt}
\renewcommand{\arraystretch}{1.05}

\begin{tabularx}{\textwidth}{l*{5}{>{\centering\arraybackslash}X}}
\toprule
\rowcolor{headerblue}
Method & Venue & PSNR$\uparrow$ & SSIM$\uparrow$
& Params(M)$\downarrow$ & GFLOPs$\downarrow$ \\
\midrule

\multicolumn{6}{c}{\textit{Single-Modal Methods}} \\
\midrule
LEDNet \citep{zhou2022lednet}        & ECCV'22    & 23.64 & 0.7295 & 7.41             & 34.44 \\
Uformer \citep{UFormer}     & CVPR'22    & 23.78 & 0.6445 & 5.29             & 24.00 \\
Retinexformer \citep{Retinexformer} & ICCV'23    & 24.06 & 0.7037 & 1.61             & 34.04 \\
MambaLLIE \citep{MambaLLIE}     & NeurIPS'24 & 24.21 & 0.6935 & 4.39             & 65.51 \\
CWNet \citep{cwnet}         & ICCV'25    & 24.01 & 0.6745 & \underline{1.29} & 21.69 \\
HVI-CIDNet \citep{hvi}    & CVPR'25    & 24.18 & 0.6826 & 1.88             & \textbf{7.57} \\
LYT-NET \citep{lyt}       & SPL'25     & 24.10 & 0.6921 & \textbf{0.04}    & \underline{13.61} \\
DarkIR \citep{darkir}        & CVPR'25    & 24.13 & 0.6831 & 3.32             & 14.23 \\
SG-LLIE \citep{dong2025towards}       & CVPRW'25   & 23.88 & 0.6259 & 12.67            & 35.80 \\
ICLRNet \citep{xu2026iclr}          & AAAI'26    & 24.31 & 0.6959 & 4.47             & 31.91 \\
\midrule

\multicolumn{6}{c}{\textit{Multimodal Methods}} \\
\midrule
\mbox{Retinexformer-M} \citep{Retinexformer}
              & ICCV'23    & 25.05             & \underline{0.8260} & 1.62 & 34.42 \\
RT-X Net \citep{jha2025rt}      & ICIP'25    & 24.53             & 0.7364             & 1.64 & 992.49 \\
DarkIR-M \citep{darkir}      & CVPR'25    & 24.59             & 0.7010             & 3.32 & 14.54 \\
ICLRNet-M \citep{xu2026iclr}        & AAAI'26    & \underline{25.16} & 0.7545             & 4.47 & 36.77 \\

\rowcolor{lastgray}              
\textbf{Ours} & --         & \textbf{25.38}    & \textbf{0.8439}    & 2.93 & 30.89 \\
\bottomrule
\end{tabularx}

\end{table*}

\begin{table*}[t]
\centering

\caption{
Quantitative comparison with representative single-modal and multimodal low-light image enhancement methods on DroneVehicle. The best and second-best results are shown in bold and underlined, respectively. "-M" denotes a thermal-input variant constructed in this work. 
}
\label{tab:quantitative_comparison_dronevehicle}

\small
\setlength{\tabcolsep}{5pt}
\renewcommand{\arraystretch}{1.05}

\begin{tabularx}{\textwidth}{l*{5}{>{\centering\arraybackslash}X}}
\toprule
\rowcolor{headerblue}
Method & Venue & PSNR$\uparrow$ & SSIM$\uparrow$
& Params(M)$\downarrow$ & GFLOPs$\downarrow$ \\
\midrule

\multicolumn{6}{c}{\textit{Single-Modal Methods}} \\
\midrule
LEDNet \citep{zhou2022lednet}      & ECCV'22    & 22.11 & 0.5468 & 7.41             & 34.44 \\
Uformer \citep{UFormer}            & CVPR'22    & 22.16 & 0.5298 & 5.29             & 24.00 \\
Retinexformer \citep{Retinexformer} & ICCV'23    & 22.32 & 0.5782 & 1.61             & 34.04 \\
MambaLLIE \citep{MambaLLIE}        & NeurIPS'24 & 22.36 & 0.6152 & 4.39             & 65.51 \\
CWNet \citep{cwnet}                & ICCV'25    & 22.33 & 0.6022 & \underline{1.29} & 21.69 \\
HVI-CIDNet \citep{hvi}             & CVPR'25    & 22.34 & 0.6039 & 1.88             & \textbf{7.57} \\
LYT-NET \citep{lyt}                & SPL'25     & 22.31 & 0.5983 & \textbf{0.04}    & \underline{13.61} \\
DarkIR \citep{darkir}              & CVPR'25    & 22.38 & 0.6188 & 3.32             & 14.23 \\
SG-LLIE \citep{dong2025towards}    & CVPRW'25   & 22.39 & 0.6069 & 12.67            & 35.80 \\
ICLRNet \citep{xu2026iclr}            & AAAI'26    & 22.41 & 0.6359 & 4.47             & 31.91 \\
\midrule

\multicolumn{6}{c}{\textit{Multimodal Methods}} \\
\midrule
\mbox{Retinexformer-M} \citep{Retinexformer}
                                & ICCV'23    & 22.58             & 0.6665             & 1.62 & 34.42 \\
RT-X Net \citep{jha2025rt}        & ICIP'25    & \underline{22.78} & 0.6462             & 1.64 & 992.49 \\
DarkIR-M \citep{darkir}           & CVPR'25    & 22.47             & 0.6529             & 3.32 & 14.54 \\
ICLRNet-M \citep{xu2026iclr}          & AAAI'26    & 22.77             & \underline{0.6745} & 4.47 & 36.77 \\

\rowcolor{lastgray}
\textbf{Ours} & --
& \textbf{22.88} & \textbf{0.6899}
& 2.93 & 30.89 \\
\bottomrule
\end{tabularx}

\end{table*}

\begin{table}[t]

\caption{ Ablation study of the main components of RMGNet on UWOT. Checkmarks and crosses indicate whether each component is included. The best results are shown in bold.}
\label{tab:ablation_main_modules}

\centering
\setlength{\tabcolsep}{6pt}
\renewcommand{\arraystretch}{1.05}
\begin{tabular}{ccccc}
\toprule
Dual Priors & RAGCM & MSERM & PSNR$\uparrow$ & SSIM$\uparrow$ \\
\midrule
$\times$ & $\times$ & $\times$ & 24.34 & 0.8299 \\
$\checkmark$ & $\times$ & $\times$ & 24.39 & 0.8349 \\
$\times$ & $\checkmark$ & $\times$ & 24.43 & 0.8344 \\
$\times$ & $\times$ & $\checkmark$ & 24.41 & 0.8301 \\
$\checkmark$ & $\checkmark$ & $\times$ & 24.60 & 0.8323 \\
$\checkmark$ & $\times$ & $\checkmark$ & 24.48 & 0.8338 \\
$\times$ & $\checkmark$ & $\checkmark$ & 24.89 & 0.8362 \\
$\checkmark$ & $\checkmark$ & $\checkmark$
& \textbf{25.38} & \textbf{0.8439} \\
\bottomrule
\end{tabular}

\end{table}

\begin{table}[t]
\centering

\caption{
    Ablation study of the three enhancement candidates in MSERM on UWOT. A checkmark indicates that the corresponding candidate is enabled. The best results are shown in bold.
}
\label{tab:ablation_candidates}

\small
\setlength{\tabcolsep}{4pt}
\renewcommand{\arraystretch}{1.0}
\begin{tabular}{ccccc}
\toprule
Preserve & Relight & IRComp & PSNR$\uparrow$ & SSIM$\uparrow$ \\
\midrule
$\checkmark$ &  &  & 24.63 & 0.8322 \\
& $\checkmark$ &  & 24.66 & 0.8362 \\
&  & $\checkmark$ & 24.67 & 0.8367 \\
$\checkmark$ & $\checkmark$ &  & 24.72 & 0.8344 \\
$\checkmark$ &  & $\checkmark$ & 24.86 & 0.8384 \\
& $\checkmark$ & $\checkmark$ & 24.98 & 0.8398 \\
$\checkmark$ & $\checkmark$ & $\checkmark$
& \textbf{25.38} & \textbf{0.8439} \\
\bottomrule
\end{tabular}

\end{table}

\begin{figure}[pos=t]
    \centering
    \includegraphics[width=\columnwidth]{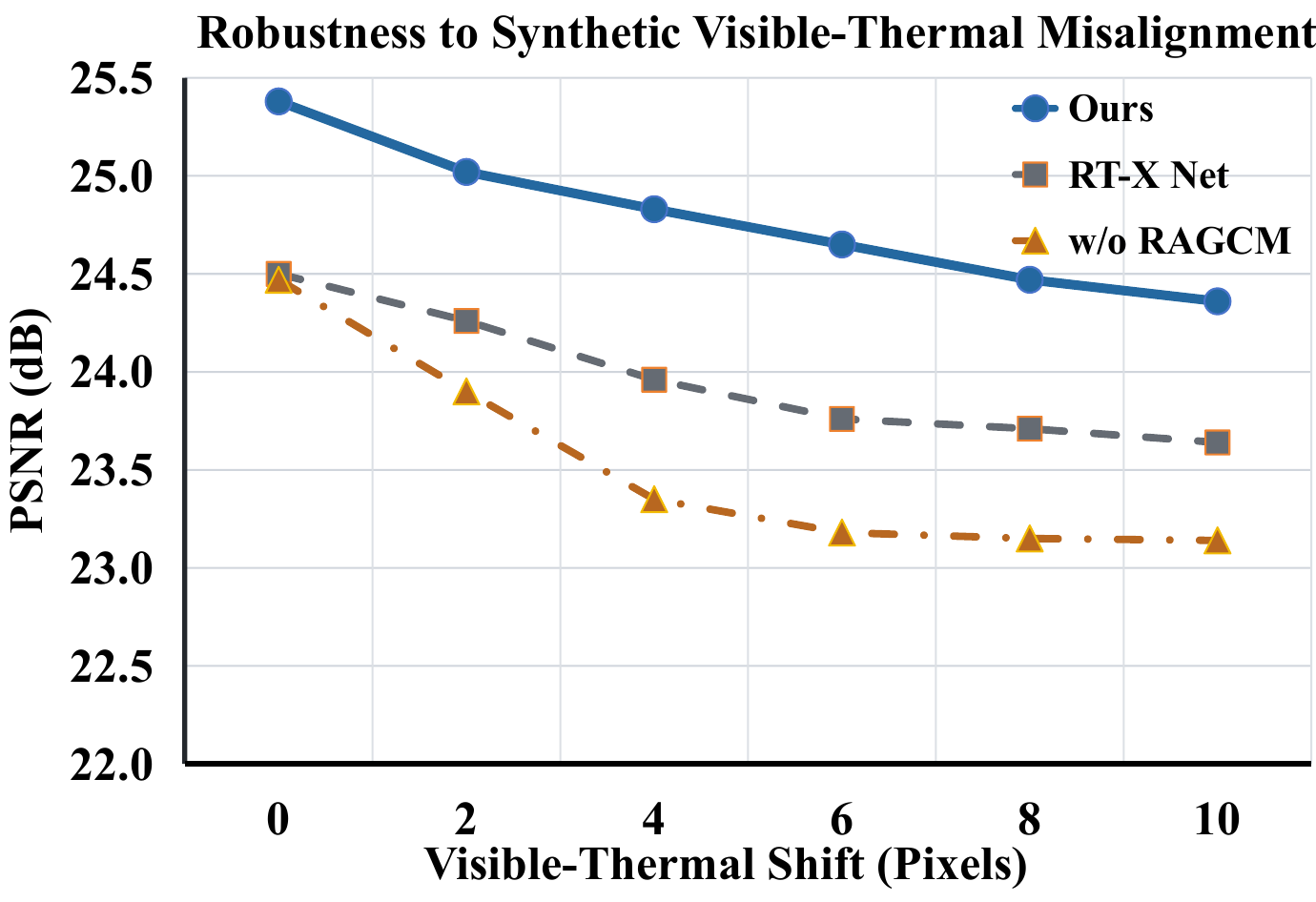}
    \caption{
    Robustness to additional synthetic optical-thermal misalignment. PSNR is evaluated at shift magnitudes of 0, 2, 4, 6, 8, and 10 pixels for the complete RMGNet, RT-X Net, and a variant of RMGNet without RAGCM. Zero shift denotes the original input pairs. Although reconstruction quality decreases with increasing displacement, the complete RMGNet maintains the highest PSNR across the evaluated range. The comparison with the variant without RAGCM illustrates the contribution of guidance calibration under the tested spatial perturbations.
}
\label{fig:misalignment_robustness}
\end{figure}

\begin{table}[t]
\centering

\caption{
    Ablation study of the training objective on UWOT. Checkmarks indicate the active loss components in each configuration. The best results are shown in bold.
}
\label{tab:ablation_losses}

\setlength{\tabcolsep}{6pt} 
\renewcommand{\arraystretch}{1.05}
\begin{tabular}{cccccc}
\toprule
${L}_{L_1}$ & $L_{\mathrm{perc}}$ & $L_{\mathrm{smooth}}$ & $L_{\mathrm{heur}}$
& PSNR$\uparrow$ & SSIM$\uparrow$ \\
\midrule
$\checkmark$ &  &  &  & 24.62 & 0.8277 \\
$\checkmark$ & $\checkmark$ &  &  & 24.80 & 0.8320 \\
$\checkmark$ & $\checkmark$ & $\checkmark$ &  & 25.13 & 0.8354 \\
$\checkmark$ & $\checkmark$ & $\checkmark$ & $\checkmark$
& \textbf{25.38} & \textbf{0.8439} \\
\bottomrule
\end{tabular}

\end{table}

We implement RMGNet in PyTorch and train it on an NVIDIA RTX PRO 6000 GPU with 96\,GB of memory. The base feature width is 32. The encoder uses channel widths of 32, 64, and 128, followed by a 256-channel bottleneck. We optimize the network using AdamW for 150 epochs with a batch size of 48 and randomly cropped $256 \times 256$ patches \citep{loshchilov2019adamw}. The initial learning rate is $2 \times 10^{-4}$, and cosine annealing reduces it to a minimum of $2 \times 10^{-6}$. Automatic mixed precision and gradient clipping are used during training.

Geometric augmentations, including random cropping, flipping, and rotation, are applied synchronously to the low-light RGB image, thermal image, and normal-light reference. This preserves their existing spatial correspondence during augmentation. For candidate refinement, the learnable scales $\gamma_p$, $\gamma_g$, $\gamma_d$, and $\gamma_i$ are initialized to 0.04, 0.10, 0.05, and 0.06, respectively. During optimization, $\gamma_p$, $\gamma_d$, and $\gamma_i$ are constrained to $[0,0.2]$, while $\gamma_g$ is constrained to $[0.01,1.0]$.

We follow the UWOT partition described in Section 4, using 4,000 training, 500 validation, and 500 test triplets. Model selection is performed on the validation set, and final results are reported on the test set. 

\noindent\textbf{Evaluation Metrics.}
Restoration performance is evaluated using peak signal-to-noise ratio (PSNR) and structural similarity index measure (SSIM), where higher values indicate better pixel fidelity and structural consistency. Model complexity and computational cost are assessed using the number of parameters (Params) and floating-point operations (GFLOPs), respectively.

Downstream segmentation is evaluated using foreground-averaged intersection over union ($\mathrm{IoU}_{\mathrm{fg}}$), foreground-averaged Dice ($\mathrm{Dice}_{\mathrm{fg}}$), and overall pixel agreement (ACC). Per-class IoU and Dice are computed from confusion counts accumulated over the entire evaluation set and averaged across the four foreground classes, excluding classes absent from both predictions and references. ACC includes all five classes. These metrics measure consistency with normal-light SAM 3 pseudo-labels rather than manually annotated ground truth.

Downstream vehicle detection is evaluated using $\mathrm{mAP}_{50}$, $\mathrm{mAP}_{50:95}$, and recall. $\mathrm{mAP}_{50}$ is calculated at an intersection-over-union threshold of 0.50, whereas $\mathrm{mAP}_{50:95}$ averages results over thresholds from 0.50 to 0.95 in increments of 0.05.

\noindent\textbf{Experimental settings on DroneVehicle.}
A second evaluation is conducted on a DroneVehicle-derived synthetic low-light benchmark containing 2,000 samples, with 1,600 used for training, 200 for validation, and 200 for testing. Low-light RGB inputs are generated from daytime visible images using the degradation process described in Section 4.2, while the corresponding thermal observations are retained. No dense optical–thermal registration is applied during benchmark construction. This setting evaluates restoration under the adopted synthetic degradation and residual cross-modal misalignment.

All compared methods are retrained on the same 1,600 training samples, with model selection performed on the shared validation set and final results reported on the shared test set.
Single-modal methods receive only the low-light RGB images, while multimodal methods additionally receive the corresponding thermal images.
All methods use the normal-light RGB images as restoration targets, and enhancement quality is evaluated using PSNR and SSIM under consistent preprocessing and metric computation settings.

\subsection{Baseline Construction and Comparison Protocol}
\label{sec:baseline_protocol}

We compare RMGNet with single-modal methods, including LEDNet~\citep{zhou2022lednet}, Uformer~\citep{UFormer}, Retinexformer~\citep{Retinexformer}, MambaLLIE~\citep{MambaLLIE}, CWNet~\citep{cwnet}, HVI-CIDNet~\citep{hvi}, LYT-NET~\citep{lyt}, DarkIR~\citep{darkir}, SG-LLIE~\citep{dong2025towards}, and ICLRNet~\citep{xu2026iclr}. For multimodal comparison, we include RT-X Net~\citep{jha2025rt} and construct thermal-input variants of DarkIR, ICLRNet, and Retinexformer, denoted by DarkIR-M, ICLRNet-M, and Retinexformer-M, respectively.

For each ``-M'' variant, the three-channel low-light RGB image and single-channel thermal image are concatenated along the channel dimension:
\begin{equation}
I_{\mathrm{multi}}
=
\operatorname{Concat}
\left(I_{\mathrm{rgb}}, I_{\mathrm{ir}}\right).
\label{eq:baseline_multimodal_input}
\end{equation}
The resulting four-channel input is processed by modifying the first input projection from three to four input channels. The subsequent backbone blocks, feature dimensions, restoration heads, and output layers are retained. No additional thermal encoder, cross-modal attention, feature-alignment module, or reliability-estimation component is introduced. These variants provide controlled comparisons of direct thermal-input concatenation.

Each ``-M'' variant retains the optimizer, learning-rate schedule, and training objective of its corresponding visible-only baseline. In particular, DarkIR-M, ICLRNet-M, and Retinexformer-M use the original losses of their respective models. The alignment, confidence, routing, and correlation losses developed for RMGNet are not applied to the competing methods.

For UWOT and DroneVehicle, all compared methods are retrained separately on the corresponding training sets. Within each dataset, all methods use the same training, validation, and test partitions, with consistent data preprocessing and evaluation procedures. Geometric augmentations are applied synchronously to the available inputs and reference images. Model selection uses the corresponding validation set, and final results are reported on the corresponding test set. The paired comparisons between each visible-only baseline and its multimodal variant assess the effect of introducing thermal information through input-level concatenation.

\subsection{Quantitative Analysis}
\label{sec:quantitative_analysis}

Table ~\ref{tab:quantitative_comparison} reports reconstruction performance on UWOT. RMGNet achieves 25.38 dB PSNR and 0.8439 SSIM, the highest scores among the evaluated methods. Its PSNR exceeds that of ICLRNet by 1.07 dB, while its SSIM exceeds that of LEDNet by 0.1144; these are the strongest visible-only baselines for the respective metrics. Compared with RT-X Net, the gains are 0.85 dB and 0.1075. These results indicate closer reconstruction agreement with the reference images under the UWOT evaluation protocol.

The matched thermal-input variants provide more direct evidence of the usefulness of auxiliary thermal observations. Relative to their visible-only counterparts, DarkIR-M, ICLRNet-M, and Retinexformer-M improve PSNR by 0.46, 0.85, and 0.99 dB, respectively. RMGNet further exceeds ICLRNet-M by 0.22 dB in PSNR and Retinexformer-M by 0.0179 in SSIM. These comparisons support the use of thermal information, while the component ablations examine the contribution of the proposed guidance and restoration design.

Table~\ref{tab:quantitative_comparison_dronevehicle} reports the quantitative comparison on the DroneVehicle-derived low-light enhancement benchmark. RMGNet achieves the highest PSNR and SSIM among the evaluated methods, reaching 22.88 dB and 0.6899, respectively. Compared with RT-X Net, which achieves the highest competing PSNR of 22.78 dB, RMGNet improves PSNR by 0.10 dB. It also exceeds the highest competing SSIM of 0.6745, obtained by ICLRNet-M, by 0.0154. These results provide additional evidence of the effectiveness of RMGNet on synthesized low-light UAV imagery with thermal guidance. All methods are retrained and evaluated using the same data partitions and evaluation protocol.

RMGNet contains 2.93 million parameters and requires 30.89 GFLOPs according to the reported complexity measurements. Its computational cost is approximately 96.9\% lower than the 992.49 GFLOPs reported for RT-X Net. It also uses fewer parameters and fewer GFLOPs than ICLRNet-M, which requires 4.47 million parameters and 36.77 GFLOPs. Together with the reconstruction results, these measurements indicate a favorable balance between restoration quality and model complexity.

\subsection{Qualitative Analysis}
\label{sec:qualitative_analysis}

We qualitatively compare RMGNet with Retinexformer, ICLRNet, and RT-X Net on representative test samples from UWOT and DroneVehicle. The comparisons examine overall illumination recovery, local contrast, and structural detail preservation, with normal-light RGB images serving as references. The full-scene results show the recovered appearance, while the enlarged regions facilitate closer inspection of local boundaries and fine structures.

On UWOT, Figure~\ref{fig:qualitative_comparison} presents two representative examples in which the low-light RGB inputs exhibit poor visibility and severely weakened structural details. The thermal inputs, shown in the insets outlined in red, retain recognizable road markings and scene contours, providing complementary structural information. Although the compared methods improve scene visibility, their results differ in local contrast and the clarity of road markings. In the yellow-boxed region, RMGNet recovers more distinguishable lane markings with clearer separation from the surrounding road surface, while several compared results exhibit faint or blurred markings. The green-boxed region further highlights directional road markings, whose arrow shapes are difficult to distinguish in the low-light input. RMGNet restores recognizable marking shapes and improves their local contrast, showing closer visual agreement with the normal-light reference in this region. Nevertheless, some marking edges and fine textures remain softer than those in the references.

\begin{figure*}[pos=t]
    \centering
    \includegraphics[width=\textwidth]
    {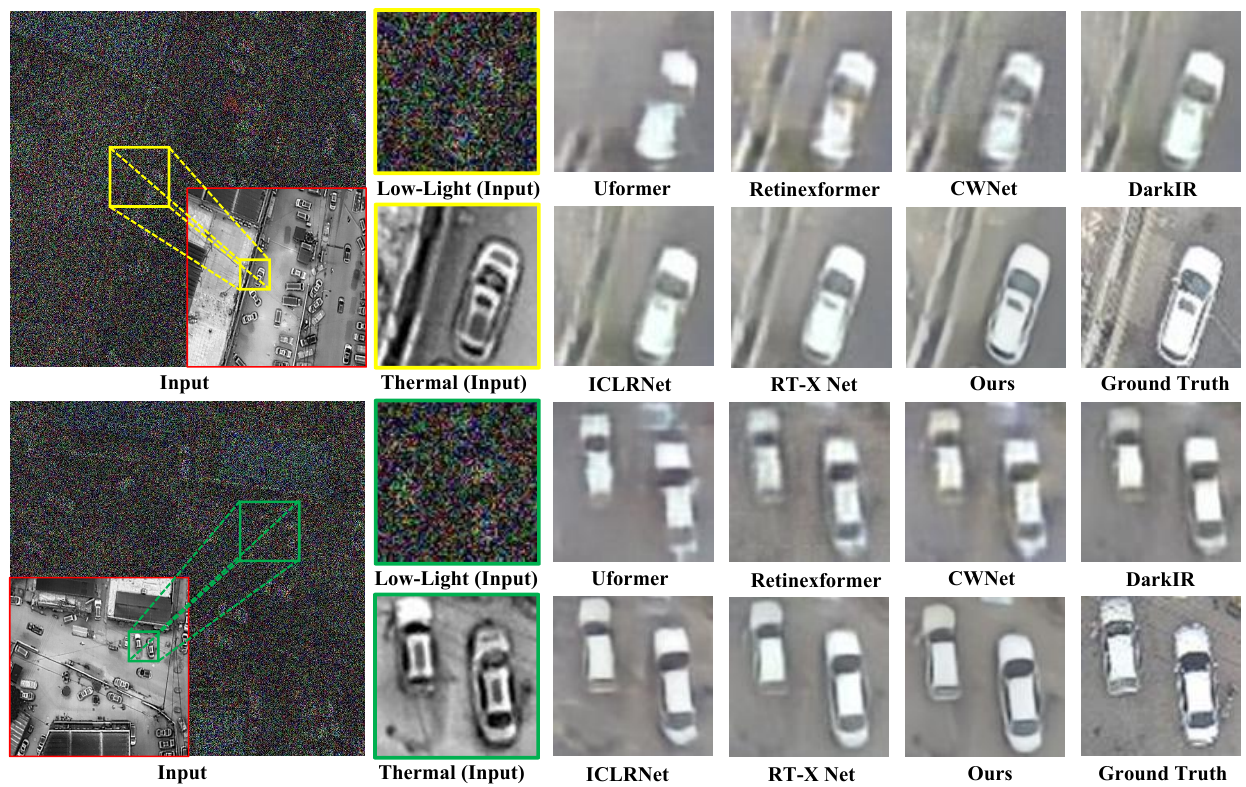}
    \caption{
        Qualitative comparison on DroneVehicle test samples. Red-bordered insets show thermal inputs; yellow and green boxes mark the regions enlarged on the right.
    }
\label{fig:qualitative_dronevehicle}
\end{figure*}

On DroneVehicle, Figure~\ref{fig:qualitative_dronevehicle} presents two representative examples of vehicle restoration under severe low-light degradation. The RGB inputs exhibit strong noise and extremely limited visibility, making vehicle shapes difficult to distinguish. In contrast, the paired thermal inputs retain recognizable vehicle contours and provide complementary structural cues. Although the compared methods improve visibility, their outputs differ in vehicle boundary definition, local contrast, and detail preservation. RMGNet restores recognizable vehicle shapes while maintaining their spatial arrangement in agreement with the normal-light references.

The enlarged regions further illustrate these differences. The yellow-boxed region in the upper example focuses on a single vehicle. RMGNet recovers a coherent vehicle outline and distinguishable window regions, whereas several compared results exhibit blurred contours or less clearly defined internal structure. The green-boxed region in the lower example contains two adjacent vehicles. RMGNet preserves their distinct shapes and separation from the surrounding road surface, producing an appearance that more closely resembles the normal-light reference. Nevertheless, some fine vehicle details and road-surface textures remain smoother and less clearly resolved than those in the references, indicating that recognizable object structure can be recovered without fully restoring fine-scale appearance.

\subsection{Analysis of Learned Confidence and Routing}
\label{sec:confidence_routing_analysis}

\begin{figure}[pos=t]
    \centering
    \includegraphics[width=\linewidth]{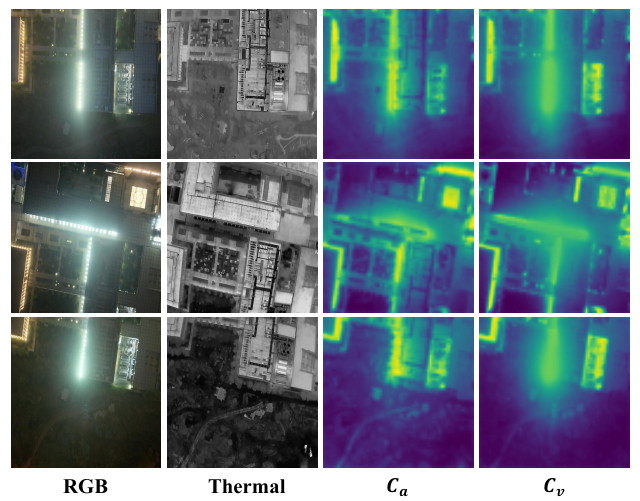}
    \caption{
        Visualization of the learned confidence maps. Each map is independently normalized, with yellow and dark purple indicating relatively high and low values, respectively; colors are not directly comparable across maps.
    }
    \label{fig:confidence_maps}
\end{figure}

\begin{figure}[pos=t]
    \centering
    \includegraphics[width=\linewidth]{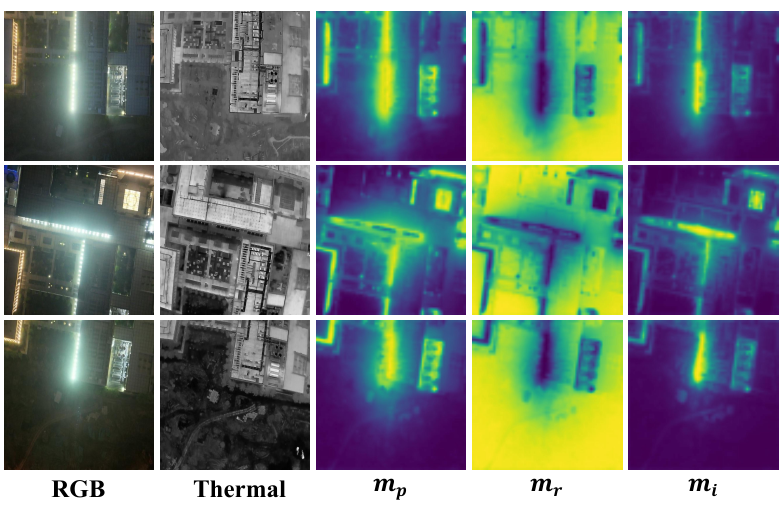}
    \caption{
        Visualization of the learned routing maps. Each map is independently normalized to show spatial variation; colors do not indicate comparable absolute weights across candidates.
    }
    \label{fig:routing_maps}
\end{figure}

To examine how RMGNet responds to spatially varying input quality, we separately visualize the learned confidence maps in Figure~\ref{fig:confidence_maps} and the routing maps in Figure~\ref{fig:routing_maps}. Both figures present the same three input pairs in the same row order, allowing the confidence estimates and restoration preferences to be examined at corresponding image locations. The confidence maps characterize the estimated usability of the available evidence, while the routing maps determine the spatial combination of the enhancement candidates.

Figure~\ref{fig:confidence_maps} shows the alignment confidence $c_a$ and visible confidence $c_v$ predicted by RAGCM. The alignment confidence estimates the local reliability of thermal guidance after feature alignment, whereas the visible confidence characterizes the usability of the remaining visible information under low-light degradation. These two quantities address different aspects of the restoration problem: the usefulness of a thermal feature depends on its correspondence with the visible image, while visible evidence may be weakened by insufficient illumination and noise. Consequently, the two confidence maps are estimated separately and are not constrained to form complementary probabilities.

In the displayed examples, $c_a$ exhibits localized variations around elongated structures and rectilinear boundaries, whereas $c_v$ tends to show smoother responses over broader regions associated with illuminated content. For example, the prominent vertical structures in the first and third rows and the horizontal structures in the second row are reflected in both maps, but their response distributions differ around the surrounding regions. Large dark areas generally exhibit lower relative responses in $c_v$ than the illuminated structures within the same map. These observations are consistent with the distinct roles of cross-modal guidance reliability and visible-evidence usability, while the maps themselves remain learned estimates rather than direct measurements of alignment correctness.

Figure ~\ref{fig:routing_maps} shows the weights assigned to preservation, relighting, and thermal-compensation candidates. The original weights are jointly softmax-normalized at each pixel, allowing a continuous mixture of restoration operations.

The displayed preservation weights have relatively high responses along prominent linear structures and locally visible content. Relighting weights vary over darker surroundings, whereas thermal-compensation weights are more concentrated around selected boundaries. These patterns are compatible with the intended roles of the candidates and illustrate spatially varying enhancement behavior.

The routing maps weight restoration operations, while the shared reconstruction features already contain fused visible and thermal information. The thermal-compensation weight alone therefore does not quantify the thermal contribution to the final output. Its effect also depends on the guidance confidence and the candidate-generation process.

The reconstruction contribution of the associated components is assessed quantitatively by the ablations in Section 5.8.
 
\subsection{Robustness to Cross-Modal Misalignment}
\label{sec:misalignment_robustness}

Spatially displaced thermal structures can provide misleading guidance for the restoration of visible scene boundaries. We therefore examine reconstruction performance under controlled additional optical–thermal translations, focusing on the contribution of the guidance-calibration design.

Figure~\ref{fig:misalignment_robustness} presents the PSNR results under synthetic shift magnitudes of 0, 2, 4, 6, 8, and 10 pixels. The comparison includes the complete RMGNet, RT-X Net, and a variant of RMGNet without RAGCM.
The zero-shift setting corresponds to the original input pairs, which may already contain residual cross-modal misalignment; the remaining settings introduce additional spatial displacement. RT-X Net provides a comparison with an alternative thermal-guided enhancement method, while the variant without RAGCM is used to examine the effect of removing guidance calibration.

All three methods exhibit decreasing PSNR as the shift magnitude increases, indicating that spatial disagreement between the modalities affects the effectiveness of thermal-guided restoration. Nevertheless, the complete RMGNet maintains the highest PSNR throughout the evaluated range, including the largest shift of 10 pixels. Its advantage across the tested displacement levels shows that the reconstruction benefit is retained when the input correspondence is further perturbed. This observation supports the effectiveness of the complete model under the evaluated synthetic translations.

The comparison with the variant without RAGCM further illustrates the importance of guidance calibration. Removing RAGCM results in lower reconstruction quality throughout the tested range, with a clear performance gap remaining at larger displacements. This behavior is consistent with the role of RAGCM in estimating cross-modal correspondence and regulating the reliability of aligned thermal guidance. When the two modalities become spatially inconsistent, the calibration mechanism helps the restoration network account for the uncertainty of thermal information before exploiting it. The ablation thus provides evidence that guidance calibration contributes to the observed reconstruction performance under
misalignment.

The complete model nevertheless remains affected by increasing displacement, as reflected by its declining PSNR. Guidance calibration therefore mitigates the adverse effects of the tested spatial perturbations without fully eliminating their influence. Moreover, maintaining the highest absolute PSNR across the tested settings does not by itself establish the smallest decrease
relative to the zero-shift baseline. The results demonstrate the benefit of the proposed calibration and restoration design within the evaluated range of synthetic translations, while more complex geometric discrepancies require
separate investigation.

\subsection{Comparison on Downstream Tasks}
\label{sec:downstream_tasks}

\begin{figure*}[pos=t]
    \centering
    \includegraphics[width=\textwidth]
    {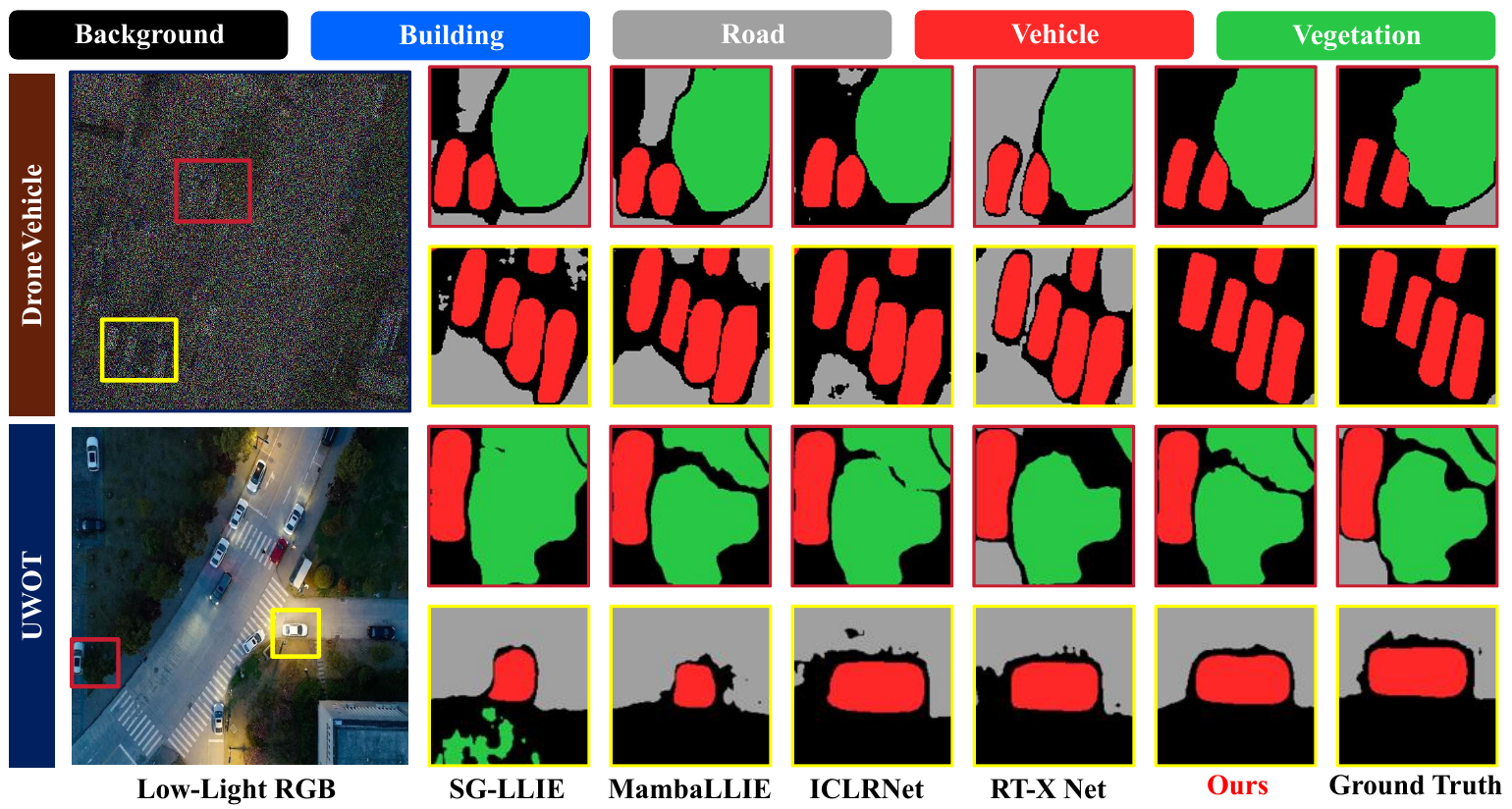}
    \caption{Qualitative comparison of segmentation consistency on DroneVehicle and UWOT. Red and yellow boxes identify the corresponding enlarged regions.}
\label{fig:downstream_seg_vis}
\end{figure*}

\begin{figure*}[pos=t]
    \centering
    \includegraphics[width=\textwidth]
    {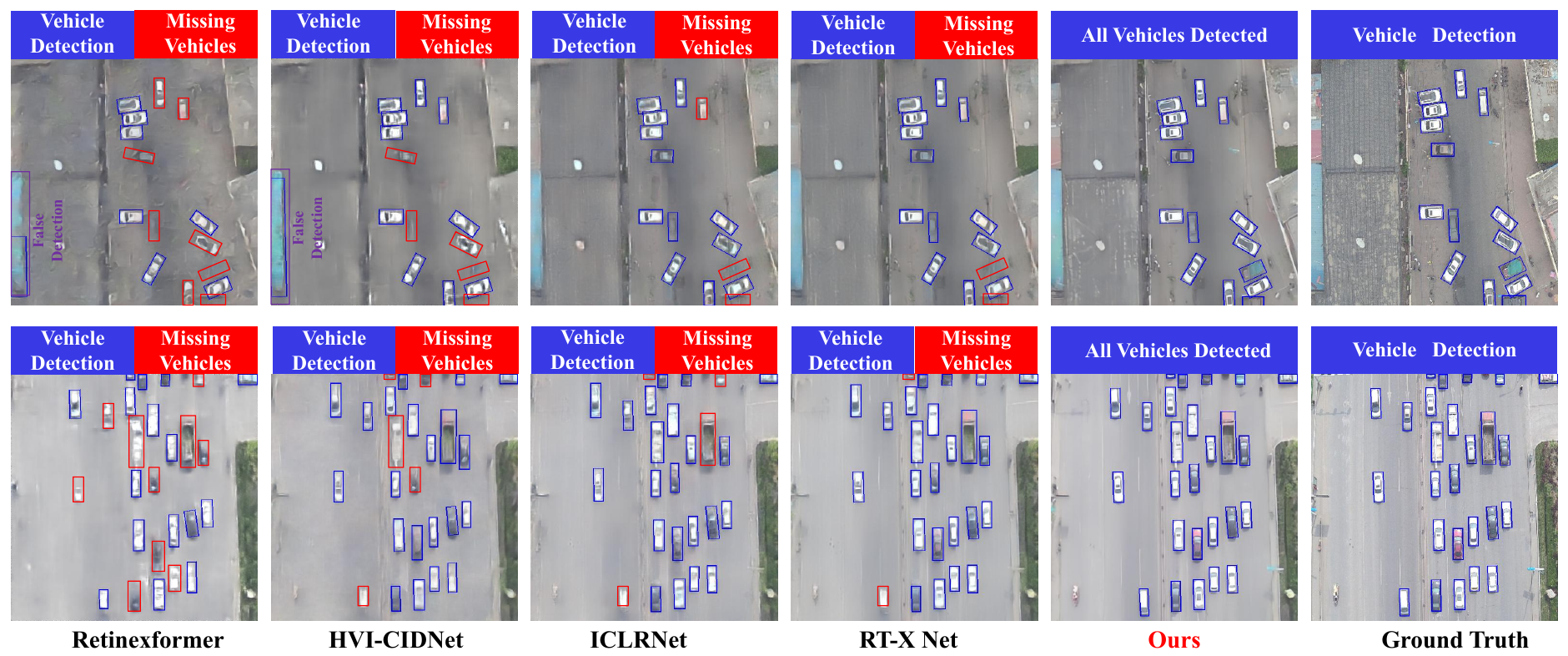}
    \caption{
        Qualitative comparison of downstream vehicle detection on two representative samples from the DroneVehicle test set. 
    }
    \label{fig:downstream_detection_visualization}
\end{figure*}

The downstream experiments examine whether the enhanced RGB images support aerial scene segmentation and vehicle detection. We conduct five-class segmentation consistency experiments using a frozen SAM 3 model \citep{sam3} on the 500 test images of UWOT and the 200 test images of DroneVehicle, together with vehicle detection experiments on DroneVehicle. These experiments complement the image reconstruction evaluation by examining the responses of downstream perception models to different enhancement results. Both downstream models receive RGB images as inputs, while thermal information contributes through the enhancement stage. Within each task and dataset, the downstream model weights, preprocessing procedures, and evaluation settings are kept identical across all compared enhancement methods. All quantitative results are reported as percentages, and performance differences are expressed in percentage points computed from the scores before rounding.

For segmentation, we employ Segment Anything Model 3 (SAM~3) with frozen pretrained weights to generate five-class label maps comprising background, building, road, vehicle, and vegetation. We use the text prompts ``building,'' ``road,'' and ``vehicle'' for the corresponding categories and merge masks prompted by ``tree,'' ``grass,'' and ``bush'' into vegetation. Instance predictions with confidence scores above 0.5 are retained, and masks are binarized at a probability threshold of 0.5. Overlapping masks are resolved by assigning each pixel to the instance with the highest confidence score, with ties broken by a fixed class order; uncovered pixels are assigned to background. Inputs are resized to $1008 \times 1008$, and output masks are restored to the original image resolution. The same checkpoint, prompts, thresholds, and processing rules are applied to all enhanced images and normal-light references. SAM~3 independently processes each enhanced image and its paired normal-light reference, with predictions on the latter serving as fixed reference pseudo-labels. 

Table~\ref{tab:downstream_seg_uwot} presents the five-class segmentation consistency results on UWOT using a frozen SAM~3 model. RMGNet achieves the highest scores across all three metrics, reaching 71.94\% $\mathrm{IoU}_{\mathrm{fg}}$, 81.91\% $\mathrm{Dice}_{\mathrm{fg}}$, and 88.91\% ACC. Compared with RT-X Net, the strongest competing method in terms of foreground IoU and Dice, RMGNet improves $\mathrm{IoU}_{\mathrm{fg}}$ and $\mathrm{Dice}_{\mathrm{fg}}$ by 1.75 and 1.56 percentage points, respectively. It also surpasses ICLRNet, which achieves the highest ACC among the competing methods, by 0.87 percentage points in ACC. Relative to ICLRNet, the strongest visible-only baseline across all three metrics, RMGNet improves $\mathrm{IoU}_{\mathrm{fg}}$ by 2.75 percentage points. These results indicate improved foreground overlap and overall pixel agreement with the normal-light SAM~3 reference predictions.

\begin{table}[t]
\centering

\caption{
    Quantitative comparison of five-class segmentation consistency on the 500 test images of UWOT using a frozen SAM~3 model. Higher values indicate greater agreement with the model-generated reference predictions. The best results are shown in bold.
}

\label{tab:downstream_seg_uwot}

\small
\setlength{\tabcolsep}{4pt}
\renewcommand{\arraystretch}{1.0}
\begin{tabular}{lccc}
\toprule
Method
& $\mathrm{IoU}_{\mathrm{fg}}(\%)\uparrow$
& $\mathrm{Dice}_{\mathrm{fg}}(\%)\uparrow$
& $\mathrm{ACC}(\%)\uparrow$ \\
\midrule
SG-LLIE
& 68.14 & 78.84 & 87.66 \\
MambaLLIE
& 68.26 & 78.28 & 87.60 \\
ICLRNet
& 69.19 & 79.46 & 88.04 \\
RT-X Net
& 70.19 & 80.35 & 87.67 \\
Ours
& \textbf{71.94}
& \textbf{81.91}
& \textbf{88.91} \\
\bottomrule
\end{tabular}

\end{table}

Table~\ref{tab:downstream_seg_dronevehicle} reports the corresponding five-class segmentation consistency results on DroneVehicle. RMGNet again achieves the highest scores among the compared methods, reaching 57.93\% $\mathrm{IoU}_{\mathrm{fg}}$, 69.71\% $\mathrm{Dice}_{\mathrm{fg}}$, and 82.40\% ACC. Compared with RT-X Net, the strongest competing method across all three metrics, RMGNet improves the corresponding scores by 3.74, 4.49, and 0.28 percentage points, respectively. It also exceeds ICLRNet, the strongest visible-only baseline in this comparison, by 5.87 percentage points in $\mathrm{IoU}_{\mathrm{fg}}$. The larger gains in foreground overlap, accompanied by a smaller improvement in overall pixel agreement, indicate that RMGNet's benefits are more pronounced for the evaluated foreground categories. Together with the UWOT results, these findings demonstrate improved consistency with normal-light reference predictions across both datasets under the same frozen SAM~3 evaluation protocol.

\begin{table}[t]
\centering

\caption{
    Quantitative comparison of five-class segmentation consistency on the 200 test images of DroneVehicle using a frozen SAM~3 model. Higher values indicate greater agreement with the model-generated reference predictions. The best results are shown in bold.
}
\label{tab:downstream_seg_dronevehicle}

\small
\setlength{\tabcolsep}{4pt}
\renewcommand{\arraystretch}{1.0}
\begin{tabular}{lccc}
\toprule
Method
& $\mathrm{IoU}_{\mathrm{fg}}(\%)\uparrow$
& $\mathrm{Dice}_{\mathrm{fg}}(\%)\uparrow$
& $\mathrm{ACC}(\%)\uparrow$ \\
\midrule
SG-LLIE
& 48.86 & 61.55 & 75.74 \\
MambaLLIE
& 50.01 & 60.80 & 79.79 \\
ICLRNet
& 52.06 & 63.20 & 82.10 \\
RT-X Net
& 54.19 & 65.22 & 82.12 \\
Ours
& \textbf{57.93}
& \textbf{69.71}
& \textbf{82.40} \\
\bottomrule
\end{tabular}

\end{table}

Figure~\ref{fig:downstream_seg_vis} presents qualitative segmentation results on representative DroneVehicle and UWOT test images. All enhanced images and their paired normal-light references are processed using the same frozen SAM~3 model. On DroneVehicle, several competing methods produce road regions around the vehicles that differ substantially from the reference predictions. In contrast, RMGNet yields a more similar distribution of vehicle, vegetation, and background regions, particularly in the yellow-boxed area containing closely spaced vehicles. On UWOT, the red-boxed region highlights differences in the vegetation outline and its separation from the neighboring vehicle, while the yellow-boxed region reveals variations in the predicted vehicle extent and the surrounding road--background boundary. RMGNet produces a vehicle mask closer in size and shape to the reference prediction and avoids the additional vegetation fragments visible in the SG-LLIE result. Although local boundary discrepancies remain, these examples complement the quantitative results by illustrating closer agreement with normal-light SAM~3 predictions in the selected regions.

For vehicle detection, we employ a fine-tuned YOLOv8m-OBB detector with the input-size parameter set to 512 \citep{jocher2023yolov8}. The detector is configured for four vehicle categories: car, freight car, truck, and bus. The van category in the original DroneVehicle dataset is excluded from the set of target categories considered in this study. Its weights remain fixed when evaluating the outputs of different enhancement methods. 

As shown in Table~\ref{tab:downstream_detection_dronevehicle}, RMGNet achieves the highest scores across all three reported detection metrics among the compared methods, reaching 64.69\% $\mathrm{mAP}_{50}$, 51.62\% $\mathrm{mAP}_{50:95}$, and 58.80\% recall. Compared with RT-X Net, the strongest competing method across these metrics, RMGNet improves the corresponding scores by 3.50, 4.84, and 1.83 percentage points, respectively. Compared with ICLRNet, the strongest visible-only baseline across the reported detection metrics, the improvements are 10.44, 8.95, and 10.45 percentage points, respectively. The improvement in $\mathrm{mAP}_{50:95}$ shows that RMGNet's advantage is maintained when detection is evaluated across multiple overlap thresholds, including stricter localization requirements. The accompanying recall improvement further shows that the average-precision gains are achieved together with higher recall under the reported evaluation protocol. These results demonstrate the utility of RMGNet-enhanced images for vehicle detection with a fixed downstream detector.

\begin{table}[t]
\centering

\caption{
    Quantitative comparison of vehicle detection results on DroneVehicle using a fixed YOLOv8m-OBB detector. All metrics are reported as percentages, with higher values indicating better performance. The best results are shown in bold.
}
\label{tab:downstream_detection_dronevehicle}

\small
\setlength{\tabcolsep}{4pt}
\renewcommand{\arraystretch}{1.0}
\begin{tabular}{lccc}
\toprule
Method
& $\mathrm{mAP}_{50}(\%)\uparrow$
& $\mathrm{mAP}_{50:95}(\%)\uparrow$
& Recall$(\%)\uparrow$ \\
\midrule
Retinexformer
& 32.80 & 25.16 & 44.28 \\
HVI-CIDNet
& 42.82 & 30.75 & 44.27 \\
ICLRNet
& 54.25 & 42.67 & 48.35 \\
RT-X Net
& 61.19 & 46.78 & 56.97 \\
Ours
& \textbf{64.69}
& \textbf{51.62}
& \textbf{58.80} \\
\bottomrule
\end{tabular}

\end{table}

Figure~\ref{fig:downstream_detection_visualization} compares vehicle detection results on two representative samples from the DroneVehicle test set. The same fixed YOLOv8m-OBB detector is applied to images enhanced by different methods and to the corresponding normal-light RGB references. The first example shows vehicles distributed along a road, whereas the second contains densely arranged parked vehicles. As shown in the figure, competing enhancement methods miss some vehicles that are detected in the normal-light RGB references, particularly small or closely spaced targets. In contrast, RMGNet successfully detects several of these missed vehicles and produces more complete detection results that are more consistent with those obtained from the reference images. These observations suggest that RMGNet better preserves the visual cues needed for vehicle detection, helping alleviate missed detections in downstream tasks.

\subsection{Ablation Study}
\label{sec:ablation_study}

We examine the contributions of the main modules, enhancement candidates, and training objective using the configurations reported in Tables~\ref{tab:ablation_main_modules}, \ref{tab:ablation_candidates}, and \ref{tab:ablation_losses}. The complete model achieves 25.38\,dB PSNR and 0.8439 SSIM.

\paragraph{Main modules.}
Table~\ref{tab:ablation_main_modules} evaluates the use of structural priors, guidance calibration, and adaptive candidate combination. Removing the dual priors, RAGCM, or MSERM reduces PSNR by 0.49, 0.90, and 0.78 dB, respectively. Among these removal configurations, excluding RAGCM causes the largest PSNR decrease, whereas excluding MSERM causes the largest SSIM decrease. The results support the combined design for reconstruction under degraded visible observations and imperfect thermal correspondence.

\paragraph{Enhancement candidates.}
Table~\ref{tab:ablation_candidates} compares the preservation, relighting, and thermal-compensation candidates individually and in combination. The single-candidate configurations obtain PSNR values between 24.63 and 24.67\,dB, below the 25.38\,dB achieved by the full model. The strongest two-candidate configuration combines relighting and thermal compensation, reaching 24.98\,dB and 0.8398. Adding preservation improves these scores by 0.40\,dB and 0.0041. Relative to the full model, removing relighting reduces PSNR by 0.52\,dB, while removing thermal compensation reduces it by 0.66\,dB. The results support the use of complementary candidate operations for regions with different restoration requirements. They also show that the bounded preservation
candidate remains useful alongside the two more active enhancement branches.

\paragraph{Training objective.}
Table~\ref{tab:ablation_losses} progressively adds perceptual, smoothness, and heuristic correlation constraints to the L1-family objective. Here, $\mathcal{L}_{L1}$ already includes reconstruction, candidate, reliability, structural, and alignment-related supervision; it is not limited to a pixel-wise reconstruction loss.
This configuration achieves 24.62\,dB and 0.8277. Adding $\mathcal{L}_{\mathrm{perc}}$ increases the scores to 24.80\,dB and 0.8320. Further introducing $\mathcal{L}_{\mathrm{smooth}}$ yields 25.13\,dB and 0.8354. The complete objective, including $\mathcal{L}_{\mathrm{heur}}$, reaches 25.38\,dB and 0.8439. The corresponding incremental PSNR improvements are 0.18, 0.33, and 0.25\,dB. These comparisons show the cumulative benefit of the additional constraints under the evaluated configurations. Because the losses are introduced sequentially, the increments describe their contributions in the presence of the previously included terms.

\section{Conclusion}
\label{sec:conclusion}

In this paper, we propose the Reliability-Aware Multi-Strategy Guidance Network (RMGNet). The proposed framework incorporates two core modules: the Reliability-Aware Guidance Calibration Module (RAGCM), which performs structural-prior-guided coarse-to-fine feature alignment and estimates the reliability of thermal guidance and the usability of visible evidence; and the Multi-Strategy Enhancement Routing Module (MSERM), which adaptively combines detail preservation, highlight-protected relighting, and confidence-gated thermal compensation to accommodate spatially varying restoration requirements. This design enables the contribution of thermal guidance and enhancement operations to vary according to the estimated local evidence quality. We also introduce UWOT, a UAV-oriented benchmark containing 5,000 triplets of low-light RGB images, normal-light RGB references, and weakly aligned thermal images. Quantitative and qualitative experiments on UWOT and a DroneVehicle-derived synthetic low-light benchmark demonstrate improved reconstruction quality over the evaluated single-modal and multimodal baselines, with clearer road markings and building outlines in the visual comparisons. Ablation studies support the combined design, while synthetic-shift experiments show that its reconstruction advantage persists under the evaluated additional translations. Downstream evaluations with fixed models further demonstrate higher vehicle-detection average precision and recall on DroneVehicle and greater consistency with normal-light reference segmentation predictions on both datasets, supporting the utility of the enhanced images for aerial road and vehicle analysis. Future research will investigate more complex cross-modal misalignment, broader real-world nighttime conditions, and deployment on resource-constrained UAV platforms.

\section*{CRediT authorship contribution statement}

\textbf{Zhicheng Zhao:} Conceptualization, Methodology, Investigation, Validation, Visualization, Writing -- original draft. \textbf{Yu Wang:} Methodology, Software, Data curation, Formal analysis, Validation. \textbf{Jiangong Xu:} Conceptualization, Supervision, Project administration, Funding acquisition, Writing -- review \& editing. \textbf{Lixiang Xu:} Investigation, Formal analysis, Writing -- review \& editing. \textbf{Chenglong Li:} Formal analysis, Validation, Writing -- review \& editing. \textbf{Jin Tang:} Supervision, Resources, Writing -- review \& editing.

\section*{Declaration of competing interest}

The authors declare that they have no known competing financial
interests or personal relationships that could have appeared to
influence the work reported in this paper.

\section*{Acknowledgments}

This work was supported in part by the National Natural Science Foundation of China (No. 62306005, 62006002, and 62076003), in part by the Natural Science Foundation of Anhui Province
(No. 2208085J18 and 2208085QF192), and in part by the Natural Science Foundation of Anhui Higher Education Institution (No. 2022AH040014).

\printcredits

\bibliographystyle{cas-model2-names}

\bibliography{cas-refs}

\end{document}